\documentclass[letterpaper]{article} 
\usepackage{main_style}
\usepackage{main_style_2}  
\usepackage{times}  
\usepackage{helvet}  
\usepackage{courier}  
\usepackage[hyphens]{url}  
\usepackage{graphicx} 
\usepackage{natbib} 
\usepackage{caption}  
\usepackage{algorithm}
\usepackage{algorithmic}

\usepackage{newfloat}
\DeclareCaptionStyle{ruled}{labelfont=normalfont,labelsep=colon,strut=off} 

\floatstyle{ruled}
\title{Repetition Makes Perfect:\\ Recurrent Graph Neural Networks Match Message-Passing Limit}
\author {
    Eran Rosenbluth, Martin Grohe
}
\affiliations{
rosenbluth@informatik.rwth-aachen.de, grohe@informatik.rwth-aachen.de\\
\vspace{0.2cm}
RWTH Aachen University
}
\begin{document}

\maketitle
\begin{abstract}
We precisely characterize the expressivity of computable Recurrent Graph Neural Networks (recurrent GNNs).
We prove that recurrent GNNs with finite-precision parameters, sum aggregation, and ReLU activation, can compute any graph algorithm that respects the natural message-passing invariance induced by the Color Refinement (or Weisfeiler-Leman) algorithm. While it is well known that the expressive power of GNNs is limited by this invariance [Morris et al., AAAI 2019; Xu et al., ICLR 2019], we establish that recurrent GNNs can actually match this limit. This is in contrast to non-recurrent GNNs, which have the power of Weisfeiler-Leman only in a very weak, "non-uniform", sense where each graph size requires a different GNN to compute with. Our construction introduces only a polynomial overhead in both time and space.

Furthermore, we show that by incorporating random initialization, for connected graphs recurrent GNNs can express all graph algorithms. In particular, any polynomial-time graph algorithm can be emulated on connected graphs in polynomial time by a recurrent GNN with random initialization.
\end{abstract}

\section{Introduction}
\subsubsection*{Graph Neural Networks}
Message-Passing Graph Neural Networks (GNNs) \cite{kipf2016semi,gilmer2017neural} is a class of graph-processing architectures commonly used in tasks of learning on graphs. As such, characterizing their expressivity is of great importance.

A GNN is a finite sequence of operations, called \emph{layers}, applied in parallel and in sync to each node. Its inputs are graphs whose nodes are assigned an initial feature-vector of certain dimension. In the broadest theoretical setting the vector is over the real numbers, however, most expressivity studies consider more restrictive domains: Some consider integers, some consider a compact segment $[a,b]\subset\Real$, and most consider a finite domain. A layer starts with constructing a \emph{message} from each neighbor, often being simply the neighbor's current value, although more sophisticated algorithms can also be used. Importantly, messages bear no identification of the sending node. An aggregation algorithm, typically dimension-wise sum; avg; or max, is then applied to the multiset of messages, producing a fixed-dimension value. Finally, a combination algorithm in the form of a \emph{Multilayer Perceptron} (MLP) is applied to the aggregation value and the node's current value, producing the node's new value. For node-level tasks, a node's value after the application of the last layer is considered the output of the GNN for that node. For graph embeddings, the nodes' final values are aggregated, and an MLP is applied to the aggregation value, producing the GNN's output for the graph.

The node-centric, node-indifferent, nature of the algorithm means every GNN can technically be applied to graphs of all sizes and GNNs are isomorphism-invariant. The MLP part of the layers gives GNNs their learnability qualities.

It is common for GNNs to use the same aggregation type in all layers. Those that use sum are denoted \emph{Sum-GNNs}. 

The expressivity boundaries of GNNs can be attributed to the combination of three factors:
\begin{itemize}
  \item [1.] The message-passing scheme. A main consequence of this is a distinguishing power that cannot exceed that of the Color Refinement algorithm - or 1-dimensional \emph{Weisfeiler-Leman} (1-WL) as it is sometimes referred to \cite{XuHLJ19,morris2019weisfeiler,aamand2022exponentially}.
  \item [2.] The fixed number of layers-executions. An obvious effect of that is a fixed information-radius for each node's computation. Another effect is impossibility to enhance the expressivity of the layers' algorithms (see next) by means of repetition.
  \item [3.] The combination and aggregation functions that make GNNs' layers. For reasons of learnability and runtime performance, these are of specific classes as mentioned above. While MLPs are universal approximators of continuous functions on compact domains \cite{hornik1989multilayer}, their expressivity is very limited for non-compact domains e.g. no MLP can approximate $\forall x\in\Nat\; x\mapsto x^2$. As for the aggregation functions, all mentioned choices potentially lose information about the multiset of neighbors' messages.
\end{itemize}

\subsubsection*{Expressivity Notions}
In more than a few studies \cite{chen2019equivalence, Grohe23}, an architecture is considered to be expressive of a function $f$ if and only if it \emph{non-uniformly} expresses $f$: For every graph size $n$ there exists a model $M_n$ of the architecture that approximates $f$ on all graphs \textbf{of that size}.
Non-uniform expressivity is not only a weak guarantee in theory, but it also has limited relevance to practice: First, non-uniform expressive GNNs that are non-polynomial in size are too large, before anything else. Second, all non-uniform expressive GNNs may succeed at inference time only when the input-graph size does not exceed the sizes of graphs in training time, otherwise they fail miserably - as implied by the theory and observed in experiments \cite{rosenbluth2023some,rosenbluthdistinguished}.

The strongest and most meaningful notion of expressivity, both to theory and practice, and the one that we use in this paper, is \emph{uniform} expressivity. A GNN architecture is said to uniformly express a function $f$ if and only if there exists (at least) one specific model of that architecture that approximates $f$ \textbf{on graphs of all sizes}. It implies that the expressing GNN computes a function that generalizes in size,  rather than computing a sort of lookup table for the finitely many graphs of a specific size. From a practical standpoint, uniform expressivity means it may be possible to train a GNN model on graphs of smaller sizes and this model will approximate the target function also on graphs of larger sizes. We are not aware of any meaningful tight bounds shown for the uniform expressivity of GNNs.
Obviously, GNNs cannot uniformly express functions that depend on an unbounded information-radius, but their limitations go beyond that: Previous works have shown that common GNN architectures cannot uniformly express basic regression and classification functions even when they are expressible by another GNN architecture and the required information-radius is only $2$ \cite{rosenbluth2023some, grohe2024targeted}.

\subsubsection*{Recurrent Graph Neural Networks}
Recurrent GNNs \cite{scarselli2008graph,GallicchioM10} are similar to GNNs in that a recurrent GNN consists of a finite sequence of layers and those comprise the same algorithms of (non-recurrent) GNN layers. However, in a recurrent GNN the sequence can be reiterated a number of times that depends on the input. Recurrent GNNs have been used successfully in practice \cite{li2016gated,selsamlearning,bresson2018residual,tonshoff2023one} and are considered promising architectures for solving various learning tasks. In terms of their uniform expressivity, recurrent GNNs are still limited by factor (1) as they still consist of local algorithms; by definition factor (2) does not apply to them; and the question is to what extent recurrence can mitigate factor (3). Recurrent MLPs with $\relu$ activation are proved to be Turing-complete \cite{siegelmann1992computational}, implying that a layer's expressivity may be increased by reiterating it. However, it has not been clear thus far if reiteration can recover the information lost in aggregation. 

A recurrent message-passing aggregate-combine architecture, 
where the nodes are aware of the graph size, was shown to be as expressive as message-passing can be \cite{pfluger2024recurrent}. However, the layers' functions there are not restricted to be computable, let alone an MLP and a common aggregation function, making the result only an upper bound for computable recurrent GNNs.
For recurrent $\sum$-aggregation GNNs, it has been proven that a single layer GNN can distinguish any two graphs distinguishable by Color Refinement \cite{bravo2024on}. However, the proof is existential, the activation function of the GNN must not be $\relu$, and most importantly, the GNN has a parameter whose value must be a real number - of infinite precision - making it incomputable.
A tight logical bound for a computable recurrent GNN architecture, named by the authors an 'R-Simple AC-GNN[F]', is proven in \cite{ahvonenlogical}.
The architecture is defined to operate only with fixed-length float values, making it limited in one aspect, and to aggregate multisets of such values always in order (e.g. ascending), making it sophisticated in another. All in all, expressivity-wise it is a strictly weaker architecture, and essentially different, than the recurrent GNNs we study. Moreover, it is characterized in different terms than the ones we use to tightly characterize our architectures.

\subsubsection*{Message-Passing Limit}
By their definition, the uniform expressivity of computable recurrent GNNs is upper-bounded by the expressivity of a general message-passing algorithm with id-less nodes that are aware of the graph-size: A single recurring algorithm operating in parallel at each node, whose input at each recurrence is the node's value and a multiset of messages from the node's neighbors, and its output is a new value and a message to send. Importantly, the same message is broadcasted to all neighbors, and messages are received with no identification of the sending node. Recurrent GNNs are then a specific case, where the recurring algorithm is an MLP composed on an aggregation of the messages, rather than a general algorithm operating on the multiset of messages.

A well-studied representation of a node defines another important upper-bound: The class of all algorithms $\CA(G,v)$, for a graph and node, that are invariant to the Color Refinement (CR) (or 1-Weisfeiler-leman) representation of the node. The Color Refinement procedure goes back to \cite{mor65,WeisfeilerL68}, see also \cite{Grohe21} and \Cref{sec:mpc}.

\subsubsection*{New Results}
\begin{figure*}[t!]
\centering
  \includegraphics[trim={0.5cm 4.5cm 0.5cm 4.5cm},clip=false,width=.98\linewidth]{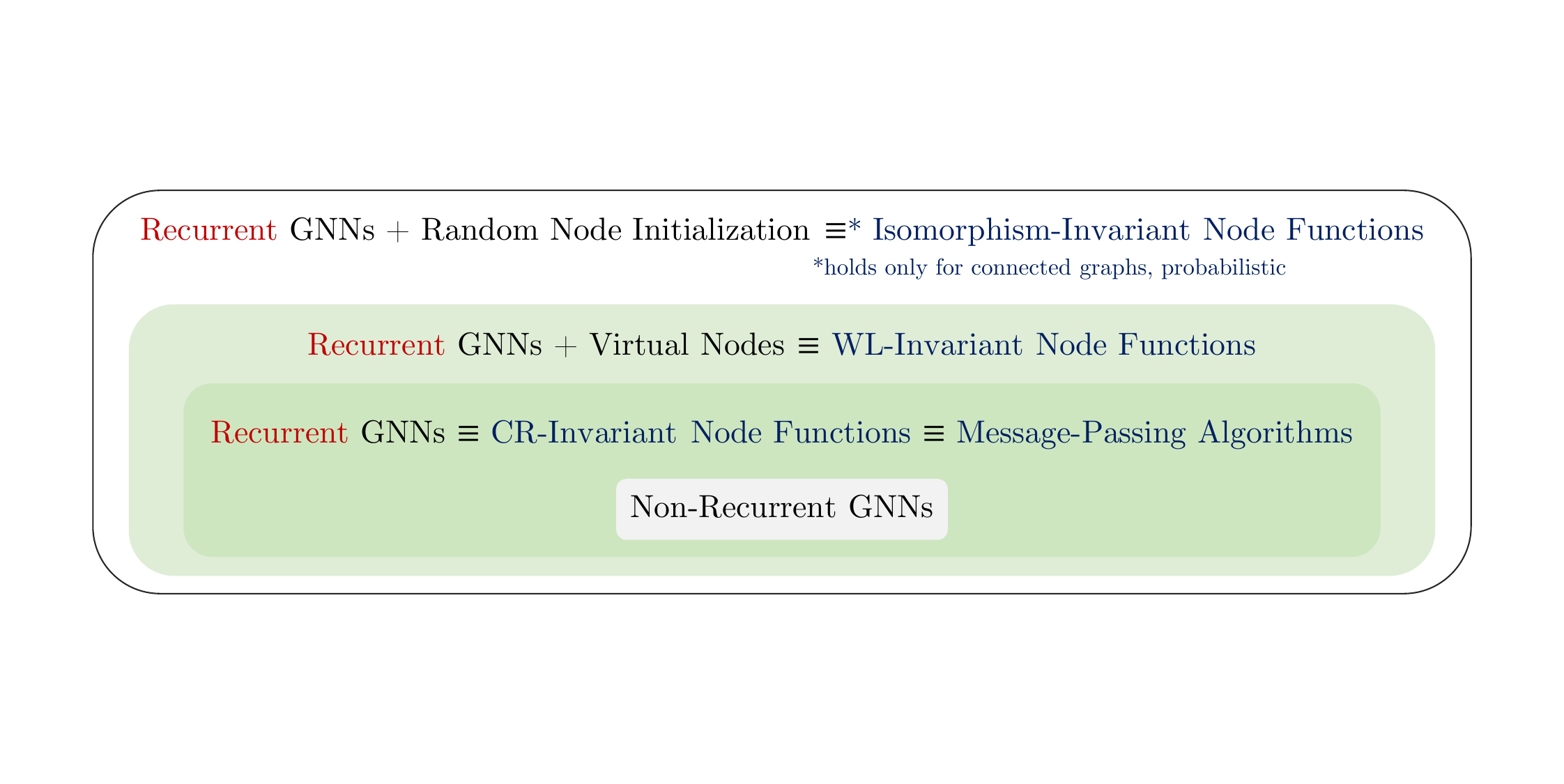}
  \caption{Uniform expressivity hierarchy. 'CR' and 'WL' are acronyms for Color Refinement and Weisfeiler-Leman. The results in this paper are the equivalencies. 'Node Functions' means computable functions $f(G,v)$ operating on a graph and node.}
  \label{fig:hierarchy}
\end{figure*}

We prove meaningful tight bounds for the uniform expressivity of computable recurrent GNNs - with finite precision weights and $\relu$ activation. All the reductions are achieved with polynomial time and space overhead.    
We assume a finite input-feature domain, and that the original feature is augmented with the graph-size value. There, we show that a recurrent single-layer sum-aggregation GNN can compute the following (see \Cref{fig:hierarchy} for an illustration):
\begin{itemize}
\item[1.] Node-level functions:
\begin{itemize}
    \item [a.] Any algorithm that is invariant to the Color Refinement (CR) value of the node. (Thm \ref{theo:main})
    \item [b.] Any message-passing algorithm. (Implied by (a))    
    \item [c.] When adding global aggregation, any algorithm that is invariant to the Weisfeiler-Leman (WL) value of the node. (Thm. \ref{thm:wlinvariant})
    \item [d.] For connected graphs, with random node initialization, any algorithm that is isomorphism-invariant (Cor. \ref{cor:main-pe}).
\end{itemize}
\item[2.] For connected graphs, any computable graph embedding that is invariant to the CR value of the graph. (Thm. \ref{thm:graphembedding})
\end{itemize}

\subsubsection*{Roadmap}
In Section \ref{sec:mpc} we describe the color of a node and its relation to the message-passing scheme, and use it to define our upper-bound function classes. 
In Section \ref{sec:main_res} we define our recurrent GNN architecture and describe the reduction from the upper-bound classes through intermediate models and down to our architecture. In Section \ref{sec:further_res} we extend our results to graph embeddings and to recurrent GNN architectures that go beyond pure message-passing.

\section{Preliminaries}\label{sec:prelim}
By $\Nat;\Rat;\Real$ we denote the natural;rational; and real numbers respectively. We define $\Ratzo\coloneqq \{q : q\in\Rat, 0\leq q\leq 1\}$.
Let $v\in\Real^d$ be a $d$-dimension vector,
we denote by $v(i)$ the value at position $i$, and by $v[a,b]\coloneqq (v_a,\ldots,v_b)$ the sub-vector from position $a$ to $b$. Let $v\in\Real^d$ and $a\in\Real$, we define $v+a\coloneqq v(1)+a,\ldots,v(d)+a$. Let $v_1,v_2\in\Real^d$ be $d$-dimension vectors, we define $v_1+v_2\coloneqq v_1(1)+v_2(1),\ldots,v_1(d)+v_2(d)$. For vectors $u\in\Real^m, v\in\Real^n$ we define 
$u,v \coloneqq u(1),\ldots,u(m),v(1),\ldots,v(n)$.
\\For a set $S$, we denote the set of all finite multisets with elements from $S$ by $\multiset{S}{*}$. We denote the set of all finite tuples with elements from $S$ by $S^*$. 
For a vector $v\in\Real^d$ we define $\dim(v)\coloneqq d$, and for a matrix $W\in\Real^{d_1\times d_2}$ we define $\dim(W)\coloneqq (d_1,d_2)$.

We define $\CB\coloneqq\{0,1\}^*$ the set of all finite binary strings, and $\CB_k\coloneqq\{0,1\}^k$ the set of all binary strings of length $k$. For $x\in \CB_k$ we define $\abs{x}\coloneqq k$.

A (vertex) \emph{featured graph} $G=\langle V(G),E(G),S,Z(G)\rangle$ is a $4$-tuple
being a graph with a \emph{feature map} ${Z(G):V(G)\rightarrow S}$, mapping each vertex to a 
value in some set $S$. Let $v\in V(G)$, we denote $Z(G)(v)$ also by $Z(G,v)$. We define the \emph{order}, or \emph{size}, of $G$, $\abs{G}\coloneqq \abs{V(G)}$. A \emph{rooted graph} is a pair $(G,v)$ where $G$ is a graph and $v\in V(G)$ is a vertex.
We denote the set of graphs featured over $S$ by $\CG_{S}$ and the set of all featured graphs by $\CG_*$.
Note that in this paper we focus on the graph domain $\CG_{\CB}$ i.e. single-dimension featured graphs where the feature is a bit-string. However, our results apply to multi-dimensional featured graphs as well: 
A node's tuple can be encoded into one dimension during pre-processing, or, alternatively, the construction in our proofs can be extended such that the initial stage of the computation includes encoding the tuple into one dimension.

We denote the set of all feature maps that map to some set $T$ by $\CZ_{T}$, and we denote the set of all feature maps by $\CZ_*$. Let $\CG\subseteq\CG_*$, a mapping $f:\CG\rightarrow \CZ_*$ to new feature maps is called a \emph{feature transformation}.

A \emph{message-passing algorithm}
\footnote{Our definition is equivalent to any distributed algorithm where the initial input includes the graph size, and the input from a node's neighbors is a multiset - with no ids or order} 
is a pair $C=(A,f)$, $A\in\CB,\ f:\CB^{2}\rightarrow \CB^2$ comprising an initial state and a computable function. It defines a feature transformation ${C:\CG_{\CB}\rightarrow \CZ_{\CB}}$ as follows: Define
\begin{center}$\forall G\in\CG_{\CB}\ \forall v\in V(G)\;
C^{(0)}(G,v)\coloneqq\te\big(A,|G|,Z(G,v)\big),\emptyset\;$\end{center}
for some encoding $\theta$ of the initial state; the graph size; and the initial feature, and $\emptyset$ denoting the empty string. Then, define  
$\forall i>0\;\ C^{(i+1)}(G,v)\coloneqq$ \begin{center}$f\Big(C^{(i)}(G,v)(1),\me(\lmulti C^{(i)}(G,w)(2):w\in N_G(v)\rmulti)\Big)$\end{center} 
for some multiset encoding $\mu$.
Define the first iteration when a 'finished' indicator turns $1$: 
$$
I_v\coloneqq\begin{cases}
			\min(i: C^{(i)}(G,v)(1)(1)=1) & \text{min exists}\\
            \infty & \text{otherwise}
		 \end{cases}
$$ 
Finally, $C(G,v)\coloneqq C^{(I_v)}(G,v)(2)$, if $I_v$ is defined.

Throughout the paper we refer to Multilayer Perceptrons (MLPs), meaning specifically $\relu$-activated MLPs, formally defined in the appendix.
A $d$ dimension \emph{recurrent MLP} $F$ is an MLP of I/O dimensions $d;d$. It defines an iterative function $f^{(t)}, t\in \Nat$ such that
      $f_F^{(0)}(x)\coloneqq x,\;\;\forall t>0\;f_F^{(t)}(x)\coloneqq f_F\big(f_F^{(t-1)}(x)\big)$.

\section{Message-Passing Information}\label{sec:mpc}
In this section, we give a precise technical description of the limits of message-passing algorithms using the \emph{Color Refinement} procedure. It aims to describe the maximal ``message-passing information'' each node can obtain. It will be convenient to refer to this message-passing information as the ``color'' of a node. Note, however, that the message-passing colors are complex objects, nested tuples of multisets, which will later be compactly represented by a directed acyclic graph (dag).

\begin{definition}\label{def:cr}
  Let $G\in \CG_{\CB_k}$. For every $t\ge 0$ and $v\in V(G)$, we define the \emph{message-passing color of $v$ after $t$ rounds} inductively as follows: The \emph{initial color} of $v$ is just its feature in $G$, that is, $ \mpc^{(0)}_G(v)\coloneqq Z(G,v).  $ The color of $v$ after $(t+1)$ rounds is the color of $v$ after $t$ rounds together with the multiset of colours of $v$'s neighbours, that is,
  \[\mpc^{(t+1)}_G(v)\coloneqq\big(\mpc^{(t)}_G(v),\lmulti \mpc^{(t)}_G(w)\mid w\in N_G(v)\rmulti\big).
  \]
  Moreover, we define the \emph{final color} of $v$ to be 
  \begin{center}$\mpc_G(v)\coloneqq\mpc^{2|G|}_G(v)$.\end{center}

  For all $t,n,k\in\Nat$, we let 
  \begin{center}
  $\MPC^{(t)}_{n,k}\coloneqq \big\{\mpc_G^{(t)}(v)\bigmid G\in\CG_{\CB_k}, |G|=n, v\in V(G)\big\}$,\\
        $\MPC^{(t)}\coloneqq\bigcup_{n,k\in\Nat}  \MPC^{(t)}_{n,k},\qquad    \MPC\coloneqq\bigcup_{t\in\Nat}  \MPC^{(t)}$.     
    \end{center}
\end{definition}                   
While most applications of Color Refinement are mainly interested in the partition of the vertices into color classes, for us, the actual colors carrying the message-passing information are important. If written as strings in a straightforward manner, the colors will
become exponentially large  (up to size $\Omega(n^t)$). 
We may also view the colors by trees, which are still exponentially large.
We introduce a polynomial-size dag representation for each color $c\in\MPC^{(2n)}_{n,k}$, denoted $D(c)$. For the full description of the dag construction, please refer to the appendix.

A feature transformation $F: \CG_{\CB}\rightarrow \CZ_{\CB}$ is \emph{message-passing-invariant} (for short: \emph{mp-invariant})
if for all graphs $G,H\in \CG_{\CB}$ of the same order $|G|=|H|$ and nodes $v\in V(G), w\in V(H)$, if $\mpc^{(t)}_G(v)=\mpc^{(t)}_H(w)$ for all $t\ge 1$ then $F(G,v)=F(H,w)$. By induction on the number of message-passing rounds, every feature transformation computed by a message-passing algorithm is mp-invariant. The converse is implied by the fact that, clearly, $D(\mpc_G(v))$ can be constructed by a message-passing algorithm, together with the following lemma which asserts that, remarkably, $D(\mpc_G(v))$ suffices to compute any mp-invariant feature transformation.
\begin{lemma}\label{lem:crinv}
  Let $F: \CG_{\CB}\rightarrow \CZ_{\CB}$ be a computable feature transformation.
  Then $F$ is mp-invariant if and only if 
  there is an algorithm that computes $F(G,v)$ from $\mpc_G(v)$. More precisely, there is
    an algorithm that, given $D(\mpc_G(v))$, computes $F(G,v)$, for all $G\in\CG_{\CB}$ and $v\in V(G)$. 
  Furthermore, if $F$ is computable in time $T(n)$ then the algorithm can be constructed to run in time $T(n)+\poly(n)$, and conversely, if the algorithm runs in time $T(n)$ then $F$ is computable in time $T(n)+\poly(n)$.
\end{lemma}
The crucial step towards proving this lemma is to reconstruct a graph from the message-passing color: Given $D(\mpc_G(v))$, we can compute, in polynomial time, a graph $G'$ and a node $v'$ such that $\mpc_{G'}(v')=\mpc_G(v)$. This nontrivial result is a variant of a theorem due to \cite{Otto97}. Once we have this reconstruction, Lemma~\ref{lem:crinv} follows easily.

\section{Main Result}\label{sec:main_res}
We would like to characterize the expressivity of R-GNNs, which operate on rational numbers, in terms of computable functions i.e. algorithms that operate on bit-strings. We use two encodings of the latter representation by the former: Rational Quaternary and Rational Binary, the reasons for which reside in the proof of \Cref{lem:smpgc_to_rgnn}. Let 
$$\RQ\coloneqq\{\Sigma^k_{i=1}a_i4^{-i} : \forall j\in[k]\;a_j\in\{1,3\},k\in\Nat\}$$
$$\RB\coloneqq\{\Sigma^k_{i=1}a_i2^{-i} : \forall j\in[k]\;a_j\in\{0,1\},k\in\Nat\}$$ 
, then define the encoding operations
$$\rqe:\CB\rightarrow\RQ,\;\rqe(b_1,\ldots,b_k)\coloneqq \Sigma^k_{i=1}(2b_i+1)4^{-i}$$
$${\rbe:\CB\rightarrow\RB},\;\rbe(b_1,\ldots,b_k)\coloneqq \Sigma^k_{i=1}b_i2^{-i}$$
For vectors of binary strings we may use the $\rqe,\rbe$ notations to denote the element-wise encoding, that is, 
$\forall (B_1,\ldots,B_l)\in\CB^l$ $$\rqe(B_1,\ldots,B_l)\coloneqq (\rqe(B_1),\ldots,\rqe(B_l)),$$$$\rbe(B_1,\ldots,B_l)\coloneqq (\rbe(B_1),\ldots,\rbe(B_l))$$

We are now ready to define the recurrent GNN architecture that is the subject of our main result. Part of the definition is the initial input provided to it. We choose to include the maximum feature length (across all nodes) $k$ in that input, as this allows us later to construct a single GNN for all $G\in\CG_{\CB}$, as stated in \Cref{theo:main}. Alternatively, we could waive having $k$ in the input, make it instead a parameter of the architecture, and restrict the statement in \Cref{theo:main} to all $G\in\CG_{\CB_k}$.

\begin{definition}\label{def:spiral_gnn}
A \emph{Recurrent Sum-GNN} (R-GNN) ${N=(A,F)}$ of dimension $d$ is a pair comprising a constant initial-state vector $A\in\Rat^{d-3}$ and an MLP $F$ of I/O dimensions $2d;d$. Dimension $d$ is a 'computation finished' indicator, and dimension $d-1$ holds the computation result. 
It defines a feature transformation ${N:\CG_{\CB}\rightarrow \CZ_{\CB}}$ as follows: Let $G\in\CG_{\CB}$, let $k\coloneqq \max(|b| : b\in \text{img}(Z(G)))$ the maximum length over the binary-string features of the vertices in $G$, and let $v\in V(G)$.
Define the initial value of $N$ to be the rational vector comprising the graph size; max feature length; initial feature; and initial state, that is,
$$N^{(0)}(G,v)\coloneqq |G|, k, \rbe(Z(G,v)), A$$
Define the value of $N$ after $t>0$ iterations to be 
$$N^{(t)}(G,v)\coloneqq F(N^{(t-1)}(G,v),\Sigma_{w\in N_G(v)}N^{(t-1)}(G,w))$$
Define the first iteration when 'finished' turns $1$
$$
I_v\coloneqq\begin{cases}
			\min(i: N^{(i)}(G,v)(d)=1) & \text{min exists}\\
            \infty & \text{otherwise}
		 \end{cases}
$$
Then, 
$$
N(G,v)\coloneqq\begin{cases}
			\rbe^{-1}\big(N^{(I_v)}(G,v)(d-1)\big) & I_v\in\Nat\\
            \text{undefined} & \text{otherwise}
		 \end{cases}
$$
the binary string represented in rational-binary encoding at position $(d-1)$, when the 'finished' indicator turns $1$.

We define a time measure $T_N(G,v)\coloneqq I_v$, and 
$$T_N(G)\coloneqq \max(I_v : v\in V(G))$$
We say that an R-GNN $N$ \emph{uses time $T(n)$}, for a function $T:\Nat\to\Nat$, if for all graphs $G$ of order at most $n$ it holds that $T_N(G)\le T(n)$.
We define $L_N(G,v)$ to be the largest bit-length over all parameters' and neurons' values of $F$, at any point of the computation for $v$, and we define $$L_N(G)\coloneqq \Sigma_{v\in V(G)}L_N(G,v)$$
We say that an R-GNN $N$ \emph{uses space $S(n)$}, for a function $S:\Nat\to\Nat$, if for all graphs $G$ of order at most $n$ it holds that $L_N(G)\le S(n)$.
\end{definition}
Note that reaching a fixed point is not required for our results, hence it is not part of R-GNNs termination definition. However, the R-GNN we construct in the proof of \Cref{lem:smpgc_to_rgnn} does have that property i.e. 
$I_v\in\Nat\Rightarrow\forall t\geq I_v\;
			 N^{(t)}(G,v)[d-1,d] = N^{(I_v)}(G,v)[d-1,d]$, which may be useful in practice and for relation to logic.

\begin{figure*}[t!]
\vspace{0.5cm}
\centering
  \includegraphics[trim={2cm 4cm 2cm 6cm},width=\textwidth]{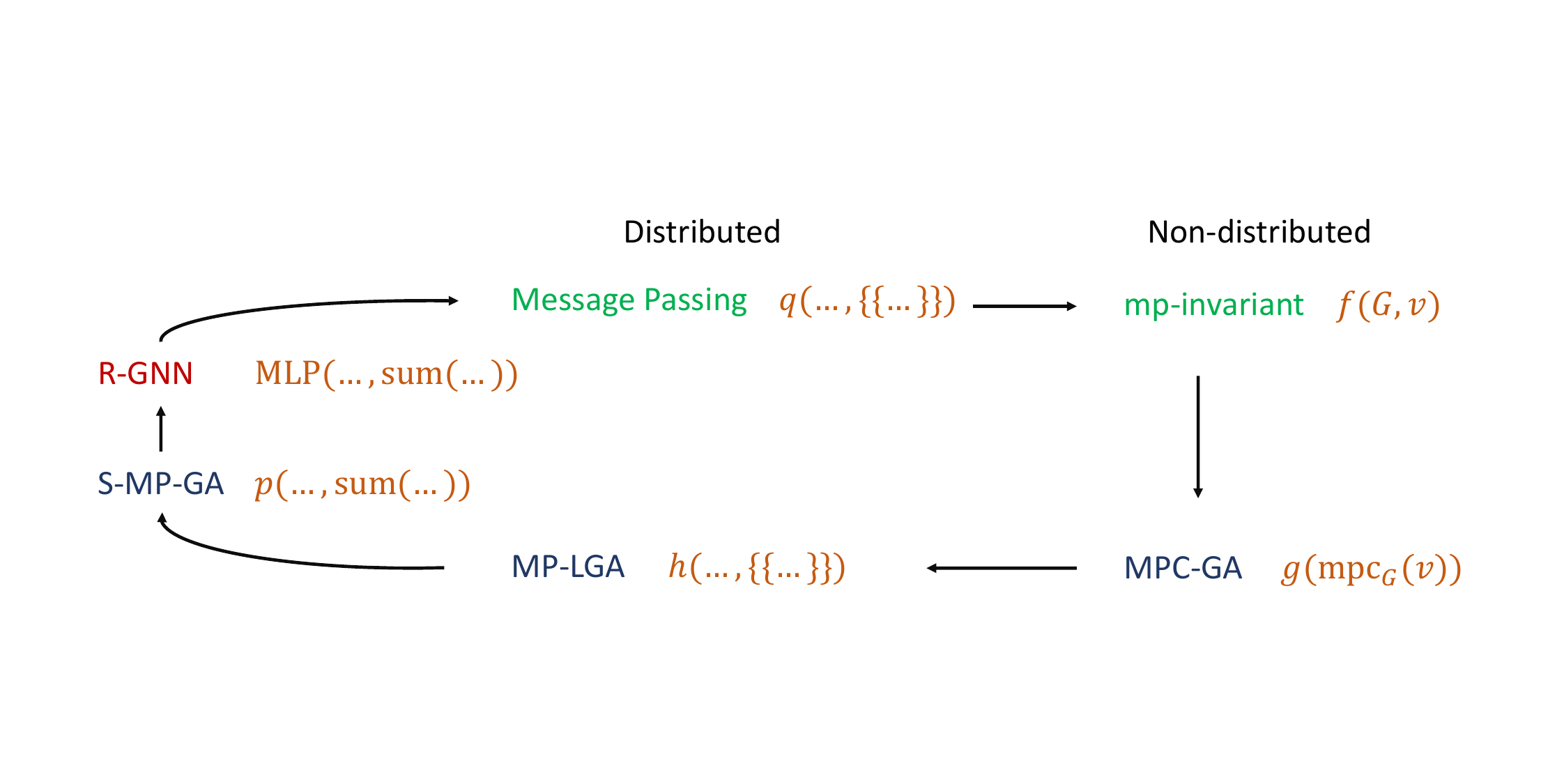}
  \caption{An overview of the reduction sequence from message-passing algorithms and mp-invariant functions to R-GNNs. Every message-passing algorithm is mp-invariant - by induction on the number of iterations. Then, starting from the mp-invariant class and moving clockwise, the reductions correspond to \Cref{lem:crinv}; Lemma \ref{lem:mpcgc_to_mpgc}; Lemma \ref{lem:mpgc_to_mpgc}; and Lemma \ref{lem:smpgc_to_rgnn}.}
  \label{fig:reductions_sequence}
\end{figure*}

By \Cref{thm:sizemust}, having the graph size in the input is a must for maximum expressivity. For recurrent GNNs with \emph{global readout} (see Section~\ref{sec:further_res}), this requirement is removed since such GNNs can compute the size.

Our main theorem refers to mp-invariant functions. However, since every message-passing algorithm is mp-invariant and since R-GNNs are specific message-passing algorithms, we have that R-GNNs are expressivity-wise equivalent also to message-passing algorithms.
\begin{theorem}\label{theo:main}
  Let $F: \CG_{\CB}\rightarrow \CZ_{\CB}$ be a computable feature transformation.
  Then $F$ is mp-invariant if and only if 
  there is an R-GNN $N$ such that \begin{center}$\forall G\in \CG_{\CB}\ \forall v\in V(G) N(G,v)=F(G)(v)$\end{center}

  Furthermore, if $F$ is computable in time $T(n)$ and space $S(n)$
  then $N$ uses time $O(T(n))+\poly(n)$ and space $O(S(n))+\poly(n)$.
\end{theorem}

\subsection{Intermediate Reductions}
Our proof of Theorem~\ref{theo:main} reduces an mp-invariant function to an R-GNN through a sequence of three intermediate computation models, see \Cref{fig:reductions_sequence} for a detailed illustration.
The models operate on bit-strings, hence we define bit-encodings for data entities that appear in the models' definitions.

We define $\de:\MPC\rightarrow \CB$ to be an encoding of space complexity $O(n^3\log n+kn)$ for all $c\in\MPC^{(t)}_{n,k}, t\leq2n$, such that all required operations can be done in polynomial time. Following the construction description of $D(\mpc_G(v))$ in \Cref{subsec:dagconstruct}, it is evident that there exists such an encoding.

We define $\me:\multiset{\CB}{*}\rightarrow \CB$ to be a multiset encoding such that the elements can be encoded and decoded in linear time by a Random Access Machine (RAM).
We define $\te:\CB^*\rightarrow \CB$ to be a tuple encoding such that the elements can be encoded and decoded in linear time by a RAM. Clearly, such encodings exist.

For $l,c,k\in\Nat$ we define $\te^{(l)}_{k,c}:(\CB_k)^l\rightarrow \CB$ to be a tuple encoding of $l$ bit-strings of length at most $k$, of space complexity $O(l(\log(c)+k)$, such that the elements can be encoded and decoded in linear time by a RAM, and, most importantly, the separation between the elements is preserved under summation of $c$ such encodings: For all $(x_1,\ldots,x_c), x_i\in(\CB_k)^l$ it holds that $$\Sigma^c_{i=1}(\te^{(l)}_{k,c}(x_i))=\te^{(l)}_{k,c}\Big(\Sigma^c_{i=1}\big(x_i(1)\big),\ldots,\Sigma^c_{i=1}\big(x_i(l)\big)\Big)$$
A straightforward encoding that reserves $\log(c2^k)$ bits for each one of the $l$ parts satisfies the requirements.

\begin{definition}\label{def:mpc_gc}
An \emph{MPC Graph Algorithm} (MPC-GA) $C=(M)$ is simply a Turing machine. It defines a feature transformation ${C:\CG_{\CB}\rightarrow \CZ_{\CB}}$ as follows:
\\Let $G\in\CG_{\CB}, v\in V(G)$, then 
$C(G,v)\coloneqq M(\de(\mpc_G(v)))$.
\end{definition}

Let $F$ be an mp-invariant function. By Lemma \ref{lem:crinv}, there exists an MPC-GA that computes $F$ with polynomial time and space overhead. Hence, the first stage in the reduction sequence in \Cref{fig:reductions_sequence} is already proven.
The next step is to translate MPC-GA - whose input is already the color of a vertex - to a distributed algorithm that has to gather that information before applying the core algorithm to it.

\begin{definition}\label{def:mp_gc}
Let $C=(A,f)$, $A\in\CB,\ f:\CB^{2}\rightarrow \CB^2$ be a pair, comprising an initial state and a computable function. It defines a feature transformation ${C:\CG_{\CB}\rightarrow \CZ_{\CB}}$ as follows: Let $G\in\CG_{\CB}$ and $v\in V(G)$, then define
$$C^{(0)}(G,v)\coloneqq\te\big(A,|G|,\de(Z(G,v))\big),\de\big(Z(G,v)\big)$$
a 2-dimension vector, the first binary string being the tuple encoding of the initial state; graph size; and encoding of the initial feature, and the second being only an encoding of the initial feature. Define  
$\;\;\;\;\forall t\geq0\;\ C^{(t+1)}(G,v)\coloneqq$ $$f\Big(C^{(t)}(G,v)(1),\me(\lmulti C^{(t)}(G,w)(2):w\in N_G(v)\rmulti)\Big)$$
the value after $t+1$ iterations.
Finally, define 
\begin{center}$C(G,v)\coloneqq C^{(2|G|+1)}(G,v)(2)$\end{center}
That is, the final output is the second output of $f$ after the $2|G|+1$ iteration.
We say that $C=(A,f)$ is a \emph{Message Passing Limited Graph Algorithm} (MP-LGA) 
if $$\forall G\in\CG_{\CB_k}\ \forall\ v\in V(G)\ \forall t\in[2|G|+1]$$
$$|\me(\lmulti C^{(t-1)}(G,w)(2):w\in N_G(v)\rmulti)|\leq O(3k|G|^4)$$
, that is, the bit-length of the multiset of neighbors' messages does not exceed $3k|G|^4$.
\end{definition}

\begin{restatable}{lemma}{lemmpcgctompgc}\label{lem:mpcgc_to_mpgc}
    Let $C=(M)$ be an MPC-GA, then there exist $A\in\CB,\ f:\CB^{2}\rightarrow \CB^2$ such that $C'=(A,f)$ is an MP-LGA and ${\forall G\in\CG_{\CB}\ \forall v\in V(G)\; C(G,v)= C'(G,v)}$. Furthermore, $C'$ incurs polynomial time and space overhead.
\end{restatable}

Next, we reduce the MP-LGA model to a model where the input from neighbors is \textbf{the sum} of the multiset of neighbors messages rather than the multiset itself. This addresses the first main obstacle of the overall reduction: To recover the information lost by the sum-aggregation and reconstruct the multiset of messages.
\begin{definition}
    Let $C=(A,f)$, $A\in\CB,\ f:\CB^{4}\rightarrow \CB^4$ be a pair, comprising an initial state and a computable function.
It defines a feature transformation ${C:\CG_{\CB_k}\rightarrow \CZ_{\CB}}$ as follows: Let $G\in\CG_{\CB_k},\ v\in V(G)$, then define
$$C^{(0)}(G,v)\coloneqq \te\big(|G|,Z(G,v),A\big),\te^{(2)}_{k,|G|}\big(1,Z(G,v)\big),0,0$$  
a vector of 4 binary strings, the first being an encoding of the graph size; initial feature; and initial state, the second being an encoding of $1$ and the initial feature, and the $3^\rd$ and $4^\th$ representing the final result and a 'finished' indicator. Define
$\;\;\;\;\forall t\geq0\;\ C^{(t+1)}(G,v)\coloneqq$ $$f\Big(C^{(t)}(G,v)(1),\;\Sigma_{w\in N(v)}C^{(t)}(G,w)(2),\;C^{(t)}(G,v)[3,4]\Big)$$
the value after $t+1$ iterations. Define 
$$
I_v\coloneqq\begin{cases}
			\min(i: C^{(i)}(G,v)(4)=1) & \text{min exists}\\
            \infty & \text{otherwise}
		 \end{cases}
$$
$$C(G,v)\coloneqq\begin{cases}
			C^{(I_v)}(G,v)(3) & I_v\in\Nat \\
            \text{undefined} & \text{otherwise}
		 \end{cases}         
$$
That is, the result is the binary string at position $3$, when the 'finished' indicator turns $1$.
We say that $C=(A,f)$ is a \emph{Sum MP Graph Algorithm} (S-MP-GA) if $\forall G\in\CG_{\CB_k}\ \forall\ v\in V(G)\ \forall t\in[2|G|+1]$ it holds that $|\Sigma_{w\in N(v)}C^{(t-1)}(G,w)(2)|\leq O(3k|G|^4)$ i.e. the bit-length of the sum of neighbors' messages is bounded by $3k|G|^4$.
\end{definition}
 Besides having a sum aggregation, an S-MP-GA differs from an MP-LGA in two technical properties:
    \begin{itemize}
        \setlength\itemsep{0.1em}
        \item [1.] A message sent by a vertex consist of two parts, 'count' and 'value', rather than one. This is in order to count the number of sending vertices, which is useful later.
        \item [2.] Two dimensions are used solely to define the final value, to make S-MP-GAs similar in that regard to R-GNNs.
    \end{itemize}
The next lemma is a key step in our sequence of reductions.

\begin{restatable}{lemma}{lemmpgctompgc}\label{lem:mpgc_to_mpgc}
    Let $C=(A,f)$ be an MP-LGA then there exist $A',f'$ such that $C'=(A',f')$ is an S-MP-GA and 
    $\;\;\forall G\in\CG_{\CB}\ \forall v\in V(G)\; C'(G,v)= C(G,v)$. Furthermore, $C'$ incurs polynomial time and space overhead.
\end{restatable}

\subsection{Reduction to R-GNN}
Finally, we reduce S-MP-GAs to R-GNNs. The essential difference between the models is the recurring algorithm: In an S-MP-GA it can be any computable function i.e. a Turing machine, while in an R-GNN it is restricted to be an MLP.
\begin{restatable}{lemma}{lemmsmpgctorgnn}\label{lem:smpgc_to_rgnn}
    Let $C=(B,h)$ be an S-MP-GA, then there exists an R-GNN $N=(A,F)$ such that $\forall G\in\CG_{\CB}\ \forall v\in V(G)\; C(G,v)= N(G,v)$. 
    Furthermore, $N$ incurs polynomial time and space overhead.
\end{restatable}
Note that an R-GNN is, in a way, an extension of a recurrent MLP to the sum-aggregation message-passing setting. In \cite{siegelmann1992computational} it is shown that recurrent MLPs are Turing-complete. Let $M$ be a Turing-machine that computes $h$, we would like to use the result in \cite{siegelmann1992computational} and emulate $M$ using the recurrent MLP in an R-GNN. Yet, this requires overcoming two significant gaps:
\begin{itemize}
    \setlength\itemsep{0.1em}
\item[1.] An encoding gap. In \cite{siegelmann1992computational} the emulation of a Turing machine is done by emulating a two-stack machine where a stack's content is always represented as a value in $\RQ$. Since $\RQ$ is not closed under summation, a naive attempt to use the sum of the neighbors' stacks directly - as input to the Turing machine emulation - is doomed to fail: The sum may be an invalid input and consequently the output will be wrong. To overcome this, we translate outgoing messages from $\RQ$ to $\RB$ - which is closed under summation, and we translate the incoming sum-of-messages back to 
$\RQ$. The translations are implemented by two dedicated recurrent sub-MLPs of $F$.
\item[2.] A synchronization gap. In an S-MP-GA computation, nodes are synchronized by definition:
The i$^\th$ recurrence's input is the the sum of results of the $i-1$ application of $h$ to each of the neighbors.
However, in an emulating R-GNN that is based on \cite{siegelmann1992computational}, every recurrence corresponds merely to a Turing machine step. As different nodes may require a different number of steps to complete the computation of a single application of $h$, when a node finishes its $i^\th$ emulation of $h$, in the $t^\th$ recurrence, its external input in recurrence $t+1$ is not necessarily the sum of its neighbors' $i^\th$ result, and it is unknown in what recurrence it will be.
To overcome this, we augment the recurrent MLP described thus far with a recurrent sub-MLP of $F$ that implements a synchronization algorithm across all nodes. That sub-MLP runs for a the same number of recurrences for all nodes, hence, in itself it is synchronized.
\end{itemize}
Overall, the recurrent MLP $F$ consists of $8$ recurrent sub-networks, each fulfilling a different task. See \Cref{fig:construction} in the appendix for an outline of $F$'s structure.
At each iteration only one sub-network, other than the synchronizer, executes its task while the others execute a computation that does not affect their main outputs.
See the appendix for further details.y

\section{Further Results}\label{sec:further_res}

Our main theorem, \Cref{theo:main}, has a number of interesting variants and implications. First, it has a version for \emph{graph embeddings} $F:\CG_{\CB}\to\CB$. 
A graph-embedding R-GNN $N=(R,M)$ comprises an R-GNN $R$ and an MLP $M$. It applies $R$, averages the nodes' values, and applies $M$. 
More formally, let $I_v$ be the last iteration of $R$ on a node $v$, then  
\begin{center}
$\forall G\in\CG_\CB\;N(G)\coloneqq \rbe^{-1}\Big(M\big(\frac{1}{|G|}\Sigma_{v\in V(G)} R^{(I_v)}(G,v)\big)\Big)$
\end{center}
We say that two graphs are \emph{mp-indistinguishable}, or indistinguishable by Color Refinement, if the same message-passing colors appear with the same multiplicities, that is, $\forall t\geq0$
\begin{center}
  $\lmulti\mpc^{(t)}_G(v)\mid v\in V(G)\rmulti= \lmulti\mpc^{(t)}_H(v)\mid v\in V(H)\rmulti$
\end{center}
Note that this implies $|G|=|H|$. A function $F:\CG_{\CB}\to\CB$ is \emph{mp-invariant} if for all mp-indistinguishable graphs $G,H$ we have $F(G)=F(H)$. Clearly, all graph embeddings computable by R-GNNs are mp-invariant. The following states the converse, which holds only for connected graphs, a limitation proved in \Cref{thm:disco}.

\begin{restatable}{theorem}{thmgraphembedding}\label{thm:graphembedding}
  Let $\mathcal{CG}_\CB\subset \CG_{\CB}$ be the subset of connected graphs in $\CG_{\CB}$, and let $F:\mathcal{CG}_\CB\to\CB$ be computable. Then, $F$ is mp-invariant if and only if there exists an R-GNN $N$ such that $\forall G\in \mathcal{CG}_\CB\; N(G)=F(G)$.
  Furthermore, if $F$ is computable in time $T(n)$ and space $S(n)$, then $N$ uses time $O(T(n))+\poly(n)$ and space $O(S(n))+\poly(n)$.
\end{restatable}

So far, R-GNNs can only compute mp-invariant functions. For connected graphs, we can break the invariance by introducing \emph{random node initialization} (RNI) \cite{AbboudCGL21}, that is, augmenting the initial feature of each node by arandom number $r\sim U(0,1)$. GNNs with RNI describe a randomized algorithm, yet this randomized algorithm, or the random variable it computes, still satisfies the usual equivariance condition i.e. its output probability-distribution is identical for every two equivariant vertices (see \cite{AbboudCGL21}). We say that an R-GNN $N$ with RNI computes a function $F:\CG_{\CB}\to\CZ_\CB$ on a graph $G\in\CG_\CB$ and node $v\in V(G)$ if and only if: $\Pr(N(G,v)=F(G,v))\ge p$ for some $\frac{1}{2}<p$, and $N(G,v)\neq F(G,v) \Leftrightarrow N(G,v)=\text{null-value}$. By repeatedly running $N$ we can boost $\Pr(N(G,v)=F(G,v))$ arbitrarily close to $1$.
\begin{restatable}{corollary}{corrandominit}\label{cor:main-pe}
  Let $\mathcal{CG}_\CB\subset \CG_{\CB}$ be the subset of connected graphs in $\CG_{\CB}$, and let $F:\mathcal{CG}_\CB\to\CZ_\CB$ be computable in time $T(n)$ and space $S(n)$. Then, there exists an R-GNN $N$ with RNI, such that $F$ is computable by $N$.
  Furthermore, $N$ uses time $O(T(n))+\poly(n)$, space $O(S(n))+\poly(n)$, and $O(n\log n)$ random bits.
\end{restatable}

Another variant of our main result concerns a common extension of GNNs which is the addition of \emph{global sum-aggregation} (a.k.a. \emph{global sum}; \emph{virtual nodes}), i.e. a sum-aggregation of the features of all nodes, as a third input to the combine MLP \cite{gilmer2017neural,BarceloKM0RS20}. Adapting Definition~\ref{def:spiral_gnn}, let $N=(A,F)$ be a GNN with global sum, then $F$ has I/O dimensions $3d;d$, and 
\begin{center}$N^{(i+1)}(G,v)\coloneqq F\big(N^{(i)}(G,v), \sum_{w\in N_G(v)}N^{(i)}(G,w),$\end{center}
\begin{center}$\sum_{w\in V(G)}N^{(i)}(G,w)\big)$\end{center}
Note that here we do not need the size of the graph as an input, since it can easily be computed using global sum. Instead of mp-invariance, R-GNNs with global sum satisfy a different invariance that we call \emph{WL-invariance}. It is based on a variant of the Color Refinement algorithm, the 1-dimensional Weisfeiler-Leman algorithm, that captures the additional global information obtained through global sum. 
See the appendix and \cite{Grohe21} for more details.

\begin{restatable}{theorem}{thmwlinvariant}\label{thm:wlinvariant}
  Let $F: \CG_{\CB}\rightarrow \CZ_{\CB}$ be a computable feature transformation.
  Then $F$ is WL-invariant if and only if 
  there is an R-GNN with global aggregation that computes $F$.
  Furthermore, if $F$ is computable in time $T(n)$ and space $S(n)$, then the R-GNN uses time $O(T(n))+\poly(n)$ and space $O(S(n))+\poly(n)$.
\end{restatable}

This theorem also has a version for graph embeddings. Here it is not restricted to connected graphs, since global aggregation provides access to all connected components.

\begin{corollary}
      Let $F:\mathcal{G}_\CB\to\CB$ be computable. Then $F$ is WL-invariant if and only if it is computable by an R-GNN.
  Furthermore, if $F$ is computable in time $T(n)$ and space $S(n)$, then the R-GNN uses time $O(T(n))+\poly(n)$ and space $O(S(n))+\poly(n)$.
\end{corollary}

\section{Concluding Remarks}
We prove that recurrent graph neural networks can emulate any message-passing algorithm, with only a polynomial time and space overhead. Thus recurrent graph neural networks are universal for message-passing algorithms, or computable mp-invariant functions. Note that our theorem is not an approximation theorem; by focusing on computable functions, we can actually construct GNNs computing the functions exactly.
By adding randomization, we can even overcome the limitation to mp-invariant functions.

Our time-complexity analysis for the reduction states "polynomial overhead", and it is not difficult to extract an upper bound of $O(n^{10}k^2)$ from our proofs.
It will be useful to know whether our reduction can be improved so to have a lower complexity, and to have a lower bound for any reduction from computable mp-invariant functions to R-GNNs. Ideally, a tight bound would be constructively proven. These remain open.

\clearpage
\section*{Acknowledgments}
During the work on this paper, Eran Rosenbluth was funded by the European Union (ERC, SymSim,
101054974) and German Research Council (DFG), RTG 2236 (UnRAVeL). Martin Grohe was funded by the European Union (ERC, SymSim,
101054974).
\bibliography{main.bib}

\clearpage

\appendix


\section{Color Refinement and Weisfeiler Leman}

\subsection*{The Represention Of a Color}\label{subsec:dagconstruct}
\begin{figure*}
    \centering
 \begin{tikzpicture}
  [ 
  thick,
  vertex/.style={fill,circle,inner sep=0pt,minimum
    size=2.5mm},
  tn/.style={fill,circle,inner sep=0pt,minimum
    size=1.5mm},
  ] 
  
  \begin{scope}[yshift=5mm]
    \node[vertex,fill=orange] (a) at (0,0) {};
    \node[vertex,fill=blue] (b) at (1.5,0) {};
    \node[vertex,fill=blue] (c) at (0,1.5) {};
    \node[vertex,fill=red] (d) at (1.5,1.5) {}; 
    \node[vertex,fill=green] (e) at (2.5,2.5) {}; 
    \draw[thick] (a) edge (b) edge (c) (d) edge (b) edge (c) edge (e);

   \end{scope}

   \node at (0.75,-0.4) {(a)};

  \begin{scope}[xshift=4cm]
        \node[tn,orange] (a1) at (0,1.5) {};
        \node[tn,orange] (a2) at (-0.5,0.75) {};
        \node[tn,orange] (a3) at (0.5,0.75) {};  
        \node[tn,orange] (a4) at (-0.75,0) {};  
        \node[tn,orange] (a5) at (-0.25,0) {}; 
        \node[tn,orange] (a6) at (0.25,0) {};  
        \node[tn,orange] (a7) at (0.75,0) {};  

        \draw[orange,->] (a1) edge (a2) edge (a3) (a2) edge (a4) edge (a5) (a3) edge (a6) edge (a7);
    \end{scope} 

    \begin{scope}[xshift=6.25cm]
        \node[tn,blue] (b1) at (0,1.5) {};
        \node[tn,blue] (b2) at (-0.5,0.75) {};
        \node[tn,blue] (b3) at (0.5,0.75) {};  
        \node[tn,blue] (b4) at (-0.75,0) {};  
        \node[tn,blue] (b5) at (-0.25,0) {}; 
        \node[tn,blue] (b6) at (0.2,0) {};  
        \node[tn,blue] (b7) at (0.5,0) {};  
        \node[tn,blue] (b8) at (0.8,0) {};  

        \draw[blue,->] (b1) edge (b2) edge (b3) (b2) edge (b4) edge (b5) (b3) edge (b6) edge (b7) edge (b8);
    \end{scope} 

    \begin{scope}[xshift=4cm,yshift=2cm]
        \node[tn,green] (e1) at (0,1.5) {};
        \node[tn,green] (e2) at (0,0.75) {};
        \node[tn,green] (e3) at (-0.5,0) {};  
        \node[tn,green] (e4) at (0,0) {};  
        \node[tn,green] (e5) at (0.5,0) {};  

        \draw[green,->] (e1) edge (e2) (e2) edge (e3) edge (e4) edge (e5);
    \end{scope} 

    \begin{scope}[xshift=6.25cm,yshift=2cm]
        \node[tn,red] (d1) at (0,1.5) {};
        \node[tn,red] (d2) at (-0.75,0.75) {};
        \node[tn,red] (d3) at (0.1,0.75) {};  
        \node[tn,red] (d4) at (0.75,0.75) {};  
        \node[tn,red] (d5) at (-1,0) {};  
        \node[tn,red] (d6) at (-0.5,0) {}; 
        \node[tn,red] (d7) at (-0.15,0) {};  
        \node[tn,red] (d8) at (0.35,0) {};  
        \node[tn,red] (d9) at (0.75,0) {};  

        \draw[red,->] (d1) edge (d2) edge (d3) edge (d4) (d2) edge (d5) edge (d6) (d3) edge (d7) edge (d8) (d4) edge (d9);
    \end{scope} 

    \path (5.125,-0.4) node {(b)};

  \begin{scope}[xshift=9cm]
        \node[tn,orange] (a1) at (0,1.5) {};
        \node[tn,orange] (a2) at (0,0.75) {};
        \node[tn,orange] (a3) at (0,0) {};  

    \footnotesize
        \draw[orange,->] (a1) edge[bend left] node[left=6pt] {$0$} (a2) edge[bend right] node[right=6pt] {$2$}(a2) (a2) edge[bend left] node[left=6pt] {$0$} (a3) edge[bend right] node[right=6pt] {$2$}(a3);

    \end{scope} 

    \begin{scope}[xshift=11.45cm]
        \node[tn,blue] (b1) at (0,1.5) {};
        \node[tn,blue] (b2) at (-0.75,0.75) {};
        \node[tn,blue] (b3) at (0.75,0.75) {};  
        \node[tn,blue] (b4) at (0,0) {};  
 
    \footnotesize
        \draw[blue,->] (b1) edge[bend left] node[above left=4pt] {$0$} (b2) edge[bend right] node[below right=4pt] {$1$}(b2) edge node[above right] {$1$}(b3)(b2) edge[bend left] node[below left=4pt] {$0$} (b4) edge[bend right] node[above right=4pt] {$2$}(b4) (b3) edge[bend left] node[above left=4pt] {$0$} (b4) edge[bend right] node[below right=4pt] {$3$}(b4);
    \end{scope} 

    \begin{scope}[xshift=9cm,yshift=2cm]
        \node[tn,green] (e1) at (0,1.5) {};
        \node[tn,green] (e4) at (-0.75,0.75) {};
       \node[tn,green] (e2) at (0.75,0.75) {};
        \node[tn,green] (e3) at (0,0) {};  

    \footnotesize
        \draw[green,->] (e1) edge node[left] {$0$} (e4) edge node[right] {$1$}(e2) (e2) edge[bend left] node[above left=4pt] {$0$} (e3) edge[bend right] node[below right=4pt] {$3$}(e3) (e4) edge[bend left] node[below left=4pt] {$0$} (e3) edge[bend right] node[above right=4pt] {$1$}(e3) ;
    \end{scope} 

    \begin{scope}[xshift=11.45cm,yshift=2cm]
        \node[tn,red] (d1) at (0,1.7) {};
        \node[tn,red] (d5) at (-1.1,0.95) {};
        \node[tn,red] (d2) at (0,0.95) {};
        \node[tn,red] (d3) at (1.1,0.95) {};  
        \node[tn,red] (d4) at (0,0) {};  
 
    \footnotesize
        \draw[red,->] (d1) edge node[above left] {$0$} (d5) edge node[left] {$2$} (d2) edge node[above right] {$1$}(d3)
        (d5) edge[bend right] node[below left] {$0$} (d4) edge node[above] {$3$}(d4)
        (d2) edge[bend right] node[left] {$0$} (d4) edge[bend left] node[right] {$2$}(d4) 
        (d3) edge[bend left] node[below] {$0$} (d4) edge node[above] {$1$}(d4);

    \end{scope} 

    \path (10.125,-0.4) node {(c)};

 \end{tikzpicture}
\caption{Two rounds of Color Refinement on a graph $G$ shown in (a). Colors can be represented as trees (b) or dags (c). The actual colors in the figure illustrate the coloring reached after two steps (they do not indicate initial features of the nodes).}
\label{fig:cr}
\end{figure*}

We introduce a succinct
representation of polynomial size, see Figure~\ref{fig:cr}(c). Every
$c\in\MPC_{n,k}^{(t)}$ will be represented by a directed acyclic graph (dag) $D(c)$
with labelled edges and labelled leaves and possibly with multiple edges between the same nodes.
$D(c)$ will have $t+1$ levels (numbered $0,\ldots,t$), each with at
most $n$ vertices. All edges will go from some level $i\ge 1$ to level
$i-1$. Edges will be labelled by natural numbers in the range
$0,\ldots,n-1$. Every vertex $v$ on a level $i$ will \emph{represent} some
color $\gamma(v)\in\MPC^{(i)}_{n,k}$ in such a way that the mapping
$\gamma$ is injective, that is, every $d\in\MPC_{n,k}^{(i)}$ is
represented by at most one node.

We say that a color $d$ is an \emph{element} of a color $c$ and write $d\in c$ if $c=(d_0,\lmulti d_1,\ldots,d_k\rmulti)$ and $d\in\{d_0,\ldots,d_k\}$.
We construct the dag $D(c)$ for $c\in\MPC_{n,k}^{(t)}$ inductively as follows. 
\begin{itemize}
    \item Level $t$ consists of a single node $u$, and we let $\gamma(u)=c$.
    \item Suppose that for some $i\in[t]$ we have defined
      level $i$ and that $C_{i}$ is the set of all colors represented by a node on level $i$. Note
      that $|C_i|\le n$, because at most $n$ distinct colors can
      appear in a graph of order $n$.

      We let $C_{i-1}$ be the set of all elements of colors in $C_i$.
      For every $d\in C_{i-1}$ we introduce a node $v_{d}$ on level $i-1$ representing $d$, that is, $\gamma(v_{d})\coloneqq d$.

    To define the edges, let $v$ be a node on level $i$ with
    $\gamma(v)=(d_0,\lmulti d_1,\ldots,d_k\rmulti)$. We add an edge
    with label $0$ from $v$ to the unique node $v'$ on level $i-1$
    with $\gamma(v')=d_0$. Moreover, for every $d\in C_{i-1}$ such
    that the multiplicity of $d$ in the multiset $\lmulti
    d_1,\ldots,d_k\rmulti$ is $\ell\ge 1$ we add an edge labelled
    $\ell$ from  $v$ to the node $v''$ on level $i-1$ with $\gamma(v'')=d$.

    \item We still need to define the labels of the leaves, that is, the nodes on level $0$. Every node
      $v$ on level $0$ represents an initial color
      $\gamma(v)\in\MPC_{n,k}^{(0)}$. Such a color is a feature of a
      graph $G\in\CG_{\CB_k}$, that is, a bitstring of length $k$, and we label $v$ by this bitstring.
\end{itemize}
This completes the description of $D(c)$. 

\begin{example}
     Consider the graph in Figure 2(a). The colors in the figure represent the colors reached after two rounds of Color Refinement. Let us consider one of the colors, say blue, and explain how the (blue) dag representing this color is constructed. The graph has no features, so all nodes get the same initial color, corresponding to the unique bottom node of the dag. Colors reached after the first round of Color Refinement correspond to the degrees of the nodes. In our blue dag, we need two of these colors, "degree 2" and "degree 3". The left node of level 1 of the blue dag represents "degree 2"; it has an edge labeled 2 to the unique node (color) on level $0$. The right node of level 1 represents "degree 3"; it has an edge labeled 3 to the unique node on level $0$. In addition, both nodes on level $1$ have an edge labeled $0$ to the unique node on level 0, indicating that in the previous round they both had the color represented by that node. On level 2 we have just one node, representing the color blue. A blue node has one neighbor of degree $2$ and one neighbor of degree $3$. Hence the top node in the dag has an edge labeled 1 to the node on level 1 representing "degree 2" and an edge labeled 1 to the node on level 1 representing "degree 3". Moreover, a blue node has degree 2 itself. Hence it has an edge labeled 0 to the node on level 1 representing "degree 2".
\end{example}

Observe that $|D(c)|\le (t+1)n$
and that for
$c,d\in\MPC_{n,k}^{(t)}$ we have $D(c)=D(d)\iff c=d$. Furthermore, it
is easy to see that given a graph $G\in\CG_{\CB_k}$ of order $n$, a
node $v\in V(G)$, and a $t\in\Nat$, we can compute
$D(\mpc^{(t)}_G(v))$ in time polynomial in $k,n,t$. 
To do this, we construct a dag simultaneously representing all colors appearing in a graph. Once we have this, we can construct a dag only representing a single color by deleting unreachable nodes. We construct the dag level by level in a bottom up fashion, maintaining pointers from the nodes of the graph to the nodes of the dag representing their color at the current level. At each level, we need to sweep through all nodes of the graph and all their incident edges, which overall requires time $O(n+m)$, where $m$ is the number of edges of the graph. Thus overall, to construct $t$ levels of the dag we need time $O(t(n+m))$. 

\subsection*{Proof of Lemma~\ref{lem:crinv}}
Before we prove the lemma, we need to make some addtional remarks on Color Refinement. Consider a graph $G$ of order $n\coloneqq |G|$. After each refinement round, we obtain a partition $\Pi_G^{(t)}$ of $V(G)$ into
\emph{color classes} $\{v\in V(G)\mid \mpc^{(t)}_G(v)=c\}$, for
$c\in\MPC^{(t)}$. As the partition $\Pi_G^{(t+1)}$ refines the
partition $\Pi^{(t)}_G$ and as $\Pi_G^{(t+1)}=\Pi_G^{(t)}$
implies $\Pi_G^{(t+2)}=\Pi_G^{(t+1)}$, for all $t$, there is a $t\le
n-1$ such that $\Pi_G^{(t+s)}=\Pi_G^{(t)}$ for all $s$. We call
$\Pi^{(t)}$ the \emph{coarsest stable partition} of $G$. However, note
that even if
$\Pi^{(t+1)}=\Pi^{(t)}$ we have $\mpc_G^{(t+1)}(v)\neq
\mpc_G^{(t)}(v)$ for all $v$ in $V(G)$, and potentially
$\mpc_G^{(t+1)}(v)$ contains relevant information not captured by
$\mpc_G^{(t)}(v)$. We will see later (Lemma~\ref{lem:no1}) that for $t\ge 2n$, this will no
longer happen. For this reason, we define the final color $\mpc_G(v)$
to be $\mpc^{(2n)}_G(v)$.

It is known that the coarsest stable partition of $G$ can be computed
in time $O(n^2\log n)$ \cite{CardonC82}. This does not mean that we can
efficiently compute the colors $\mpc_G(v)$, which may be exponentially large, but, as explained earlier, we can compute
$D(\mpc^{(t)}_G(v))$ in time polynomial in $k,n,t$.

The following lemma is just restating Lemma~\ref{lem:crinv}.
\begin{lemma}\label{alem:crinv}
  Let $F: \CG_{\CB}\rightarrow \CZ_{\CB}$ be computable.
  Then the following are equivalent.
  \begin{enumerate}
  \item $F$ is mp-invariant.
  \item There is an algorithm that computes $F(G,v)$ from the color $\mpc_G(v)$. More precisely, there is
    an algorithm that, given $D(\mpc_G(v))$ for some graph
    $G\in\CG_{\CB}$ and $v\in V(G)$, computes $F(G,v)$.
\end{enumerate}
  Furthermore, if $F$ is computable in time $T(n)$ then algorithm
  in (2) can be constructed to run in time $T(n)+\poly(n)$,
  and conversely, if the algorithm in (2) runs in time
  $T(n)$ then $F$ is computable in time $T(n)+\poly(n)$.
\end{lemma}

The proof of this lemma relies on result due to \cite{Otto97}. First of all, Otto
observed that the complete message-passing information of a graph
can be represented by a \emph{sketch} of the graph.
\footnote{In \cite{Otto97,KieferSS22}, our sketches are called
  \emph{$\mathsf{C}^2$-invariants}, refering to a logic $\mathsf{C}^2$
  that is closely related to the Color Refinement algorithm.}
We say
that an \emph{$\ell$-dimensional sketch}, for some $\ell\ge1$, is a triple $S=(A,\boldsymbol b,\boldsymbol c)$
where $A=(A_{ij})_{i,j\in[\ell]}\in\mathbb
  N^{\ell\times\ell}$, and $\boldsymbol b=(b_1,\ldots,b_\ell)\in\mathbb N_{>0}^\ell$ and $\boldsymbol c=(c_1,\ldots,c_\ell)\in\mathcal B^\ell$.
With every graph $G\in\CG_{\CB}$, we associate a sketch
$S(G)=(A,\boldsymbol b,\boldsymbol c)$ as follows: suppose that the
color classes of the coarstest stable coloring of $G$ are $V_1,\ldots,V_\ell$. Then for all
$i\in[\ell]$ we let $b_i\coloneqq|V_i|$ and $c_i\coloneqq Z(G,v)$ for
all $v\in V_i$ (using the fact that $\mpc_G(v)=\mpc_G(w)$ implies
$Z(G,v)=Z(G,w)$). Moreover, for all $i,j\in[\ell]$ we let $A_{ij}$ be
the number of neighbors that vertices in $V_i$ have in $V_j$, that is,
$A_{ij}\coloneqq |N_g(v)\cap V_j|$ for $v\in V_i$ (using the fact that
$\mpc_G(v)=\mpc_G(w)$ implies $|N_G(v)\cap V_j|=|N_G(w)\cap V_j|$
for all $j$). This definition depends on the order $V_1,\ldots,V_\ell$
in which we enumerate  the color classes. But it is easy to define a
canonical order of these classes (see \cite{Otto97,KieferSS22}), and
this is the order we always use. The following lemma shows that the
sketches capture the complete Color Refinement information. We say that Color Refinement \emph{distinguishes} two graphs, if 
\[
\lmulti\mpc^{(t)}_G(v)\mid v\in V(G)\rmulti\neq\lmulti\mpc_{G'}^{(t)}(v)\mid v\in V(G')\rmulti
\]
for some $t\in\Nat$.

\begin{lemma}[\cite{Otto97}]\label{lem:otto1}
  For all $G,H\in \CG_{\CB}$ the following are equivalent: 
    \begin{enumerate}
    \item $S(G)=S(H)$;
    \item Color Refinement does not distinguish $G$ and $H$;
    \item $\lmulti\mpc_{G}^{(t)}(v)\mid v\in V(G)\rmulti=\lmulti\mpc^{(t)}_H(w)\mid w\in V(H)\rmulti$ for some $t\ge \max\{|G|,|H|\}$.
    \end{enumerate}
\end{lemma}

The main result of \cite{Otto97} is that we can retrieve a graph from its sketch in polynomial time, that is, there is a polynomial time algorithm that, given a sketch $S$, decides if there is a graph $G\in\CG_{\CB}$ such that $S(G)=S$ and if there is computes such a graph $G$. Note that the graph the algorithm computes give a sketch $S(G)$ is not necessarily $G$, but some graph $G'$ with $S(G')=S(G)$.

Kiefer, Selman, and Schweitzer~\cite{KieferSS22} gave a simpler proof of Otto's result. The core of their proof is the following lemma, which we will also use here. 
An $\ell$-dimensional sketch $S=(A,\boldsymbol
b,\boldsymbol c)$ is \emph{realisable} if there is a graph $G$ and a
partition $V_1,\ldots,V_\ell$ of $V(G)$ such that for all $i\in[\ell]$
we have $b_i=|V_i|$ and $c_i= Z(G,v)$ for all $v\in V_i$, and for all
$i,j\in[\ell]$, each vertex $v\in V_i$ has $A_{ij}$ neighbours in
$V_j$.
We say that $G$ \emph{realises} $S$ via $V_1,\ldots,V_\ell$.

Note that
this does not mean that $S(G)=S$, because the partition
$V_1,\ldots,V_\ell$ may be a refinement of the coarsest stable
partition of $G$. For example, for every graph $G\in\CG_{\CB}$ with
vertex set $V(G)=\{v_1,\ldots,v_n\}$ we can define a trivial
$n$-dimensional sketch $S=(A,\boldsymbol b,\boldsymbol c)$ by letting
$A_{ij}\coloneqq 1$ if $v_iv_j\in E(G)$ and $A_{ij}=0$ otherwise (that
is, $A$ is the adjacency matrix of $G$), $b_i\coloneqq1$ for all $i$,
and  $c_i\coloneqq Z(G,v_i)$. It can be shown that $S(G)$ is the unique sketch of minimum dimension that is realised by $G$.

\begin{lemma}[\cite{KieferSS22}]\label{lem:kiefer}
  A sketch $S=(A,\boldsymbol
b,\boldsymbol c)$ is realisable if and only if the following
conditions are satisfied:
\begin{enumerate}
\item[(i)] $b_i\cdot A_{ii}$ is even for all $i\in[\ell]$;
\item[(ii)] $b_iA_{ij}=b_jA_{ji}$ for all $i,j\in[\ell]$.
\end{enumerate}
Furthermore, there is a polynomial time algorithm that, given a
sketch
$S$, decides if $S$ realisable and computes a graph $G$ and a
partition $V_1,\ldots,V_\ell$ of $V(G)$ realising $S$ if $S$ is
realisable.
\end{lemma}

We need versions of these lemmas on the node level. An
\emph{$\ell$-dimensional weak sketch}
is a pair $(A,\boldsymbol c)$ 
where $A=(A_{ij})_{i,j\in[\ell]}\in\mathbb
  N^{\ell\times\ell}$ and $\boldsymbol
  c=(c_1,\ldots,c_\ell)\in\mathcal B^\ell$. An 
\emph{$\ell$-dimensional weak node sketch}
is a triplet $(A,\boldsymbol c,k)$, where $(A,\boldsymbol c)$ is an
$\ell$-dimensional weak sketch and $k\in[\ell]$. 
The \emph{weak node sketch of} a connected rooted
graph $(G,v)$, 
where $G\in\CG_{\CB}$ and $v\in V(G)$, is the weak
node sketch $S_{\textup w}(G,v)$ defined as follows:
suppose that the MPC-equivalence classes of $G$ are
$V_1,\ldots,V_\ell$ (in canonical order), $v\in V_k$, and
$S(G)=(A,\boldsymbol b,\boldsymbol c)$. Then $S_{\textup
  w}(G,v)=(A,\boldsymbol c,k)$. We call this the ``weak'' sketch
because we drop the information about the sizes of the equivalence
classes, represented by the vector $\boldsymbol b$ from the sketch. The \emph{weak node sketch of} a disconnected rooted
graph $(G,v)$ is the weak node sketch $S_{\textup w}(G,v)\coloneqq S_{\textup w}(G_v,v)$, where $G_v$ is the connected component of $v$ in $G$. The idea behind this definition being that the weak node sketch should only contain information accessible to $v$ by message-passing. In the following, we always denote the connected component of a vertex $v$ in a graph $G$ by $G_v$.

A weak node sketch $S=(A,\boldsymbol c,k)$ is \emph{realisable} if there is a
rooted graph $(G,v)$ and a
partition $V_1,\ldots,V_\ell$ of $V(G_v)$ such that for all
$i\in[\ell]$, $v'\in V_i$
we have $c_i= Z(G,v')$, and for all
$i,j\in[\ell]$, each vertex $v'\in V_i$ has $A_{ij}$ neighbours in
$V_j$.
We say that $(G,v)$ \emph{realises} $S$ via $V_1,\ldots,V_\ell$.

\begin{lemma}\label{lem:no1}
  Let $G,H\in \CG_{\CB}$ be connected graphs, $v\in V(G),w\in V(H)$, and $t\ge 2\max\{|G|,|H|\}$. Then the following are equivalent:
  \begin{enumerate}
  \item there is a weak node sketch $S$ such that both $(G,v)$ and
    $(H,w)$ realise $S$;
  \item $S_{\textup w} (G,v)=S_{\textup w} (H,w)$;
  \item $\mpc^{(t)}_G(v)=\mpc^{(t)}_H(w)$;
  \item $\mpc_G^{(s)}(v)=\mpc_H^{(s)}(w)$ for all $s\ge 1$.
  \end{enumerate}
\end{lemma}

\begin{proof}
  The implications (2)$\implies$(1) and (4)$\implies$(3) are trivial.
  To prove that
  (3)$\implies$(2), assume that $\mpc_G^{(t)}(v)=\mpc_H^{(t)}(w)$
  and consider the dag $D:=D(\mpc_G^{(t)}(v))=D(\mpc_H^{(t)}(w))$.
  Recall that $D$ has $t$ levels, with edges only between successive
  levels. For $i\in[t]$, let $U^{(i)}$ be the set of all nodes on
  level $(i)$.

  Each node $u\in U^{(i)}$ represents the message-passing information
  $\mpc_G^{(i)}(v')$ of at least one node $v'\in V(G)$ and the
  message-passing information $\mpc_H^{(i)}(w')$ of at least one
  node $w'\in V(H)$. Let $V(u)\subseteq V(G)$ and $W(u)\subseteq V(H)$
  be the sets of all nodes $v'\in V(G),w'\in V(H)$, respectively,
  whose message-passing information is represented by $u$. Moreover,
  let $\mpc(u)\coloneqq\mpc_G^{(i)}(v')=\mpc_G^{(i)}(w')$ for all
  $v'\in V(u),w'\in W(u)$.

  If a node $v'\in V(G)$ is reachable from $v$ by a path of length
  at most $j$, then $v'\in V(u)$ for some $u\in U^{(i)}$ for every
  $i\le t-j$. Let $n\coloneqq\max\{|G|,|H|\}\le \lfloor{t/2}\rfloor$. 
  Since $G$ is connected, every node is
  reachable by a path of length at most $n-1$. This implies
  that for all $i\le n+1$, $\big(V(u)\big)_{u\in U^{(i)}}$ is a
  partition of $V(G)$.    Similarly, $\big(W(u)\big)_{u\in U^{(i)}}$ is a
  partition of $V(H)$. Thus
  \[
    \big\{\mpc(u) \;\big|\; u\in
    U^{(i)}\big\}
    =\big\{\mpc_G^{(i)}(v')\;\big|\; v'\in V(G)\big\}
=\]\[\big\{\mpc_H^{(i)}(w') \;\big|\; w'\in V(H)\big\}.
  \]
  Note that these are set equalities, not multiset equalities. It
  may well be that for $u\in U^{(i)}$ we have $|V(u)|\neq|W(u)|$,
  and thus the multiplicities of $\mpc(u)$ in the multisets
  $\lmulti \mpc_G^{(i)}(v')\mid v'\in V(G)\rmulti$ and $\lmulti
  \mpc_H^{(i)}(w')\mid w'\in V(H)\rmulti$ are different.
  
  As the refinement process on $G$ and $H$ stabilizes in at most $n$
  iterations, $\big(V(u)\big)_{u\in
    U^{(n)}}$ is the coarsest stable partition of $G$.
  Moreover, as 
  $\big(V(u)\big)_{u\in U^{(n+1)}}$ is still a partition of
  $V(G)$, we have $\big(V(u)\big)_{u\in
    U^{(n+1)}}=\big(V(u')\big)_{u'\in U^{(n)}}$. The corresponding
  matching between $U^{(n+1)}$ and $U^{(n)}$ is given by the edges in
  $D$ from level $n+1$ to level $n$ with
  label $0$.
  For all $u\in
  U^{(n+1)}$, $u'\in U^{(n)}$, the message passsing information
  $\mpc(u)$ tells us the number of neighbours each $v\in V(u)$ has
  in $V(u')$. Thus for all $u,u'\in U^{(n)}$ there is a number $A_{uu'}$
  such that each $v\in V(u)$ has $A_{uu'}$ neighbours in $V(u')$.
  Similarly, $\big(W(u)\big)_{u\in
    U^{(n)}}$ is the coarsest stable partition of $H$, and for all $u,u'$ each vertex in
  $W(u)$ has $A_{uu'}$ neighbours in $W(u')$; since $A_{uu'}$ only
  depends on $\mpc(u'')$ for the node $u''\in U^{(n+1)}$ connected to
  $u$ by an edge with label $0$, it is the same in both $G$ and $H$.
  
  Furthermore, for all $u\in U^{(n)}$ there is a $c_u\in\CB$ determined by
  $\mpc(u)$ such that $Z(G,v')=c_u$ for all $v'\in V(u)$ and
  $Z(H,w')=c_u$ for all $w'\in W(u)$.
  
  Letting $A\coloneqq (A_{uu'})_{u,u'\in U^{(n)}}$, $\boldsymbol
  b=(b_u)_{u\in U^{(n)}}$ with $b_u\coloneqq |V(u)|$, $\boldsymbol
  b'=(b'_u)_{u\in U^{(n)}}$ with $b'_u\coloneqq |W(u)|$, and
  $\boldsymbol c=(c_u)_{u\in U^{(i)}}$, we have $S(G)=(A,\boldsymbol
  b,\boldsymbol c)$ and $S(H)=(A,\boldsymbol
  b',\boldsymbol c)$.
  
  Moreover, tracing the unique path of edges labelled $0$ from the
  unique node $u\in U^{(t)}$ with
  $\mpc(u)=\mpc^{(t)}_G(v)=\mpc^{(t)}_H(w)$ to level $n$, we
  find a node $u'\in U^{(n)}$ such that
  $\mpc(u')=\mpc^{(n)}_G(v)=\mpc^{(n)}_H(w)$. Then we have
  \[
    S_{\textup w} (G,v)=(A,\boldsymbol c,u')=S_{\textup w} (H,w).
  \]
  
  \medskip
  It remains to prove the implication 
  (1)$\implies$(4). Assume that $S=(A,\boldsymbol c,k)$, where $A=(A_{ij})_{i,j\in[\ell]}$ and $\boldsymbol c=(c_1,\ldots,c_\ell)$ for some $\ell\in\Nat$, is a weak node
  sketch realised by $(G,v)$ via the partition $V_1,\ldots,V_\ell$ and
  realised by $(H,w)$ via the partition $W_1,\ldots,W_\ell$. 
  
  Let $s\ge 1$. It is our goal to prove that
  \begin{equation}
    \label{eq:2}
    \mpc_G^{(s)}(v)=\mpc_H^{(s)}(w).
  \end{equation}
  Let $n\coloneqq\max\{|G|,|H|\}$, and let
  $D_G\coloneqq D(\mpc^{(s+n)}_G(v))$ and $D_H\coloneqq
  D(\mpc^{(s+n)}_H(w))$.
  For every $i\in[s+n]$, let $D^{(i)}_G$ be the restriction of $D_G$
  to the first $i$ levels, and let $D_H^{(i)}$ be the restriction of
  $D_H$ to the first $i$ levels. Let $U^{(i)}_G,U^{(i)}_H$ be the sets of nodes on level $i$ in the respective graph. Every $u\in U^{(i)}_G$
  represents some color $\mpc_G(u)$ such
  that there is a $v'\in V(G)$ with $\mpc_G^{(i)}(v')=\mpc_G(u)$.
  We let $V^{(i)}_u$ be the set of all such $v'$. Then the sets $V^{(i)}_u$ for
  $u\in U^{(i)}_G$ are mutually disjoint. Arguing as above, we see
  that for $i\le s$ the sets $V^{(i)}_u$ for $u\in U^{(i)}_G$ form a partition
  of $V(G)$. Similarly, for every $u\in U^{(i)}_H$ there is some
  color $\mpc_H(u)$ and a vertex $w'\in V(H)$
  with $\mpc_H^{(i)}(w')=\mpc_H(u)$. We let $W^{(i)}_u$ be the set of
  all such $w'$. Then for $i\le s$, the sets $W^{(i)}_u$ for $u\in U^{(i)}_H$ form a partition
  of $V(G)$.
  
  By induction, we shall prove that for all $i\in[s]$, the following
  conditions are satisfied.
  \begin{enumerate}
  \item[(i)] $D^{(i)}_G=D^{(i)}_H$.\\
    In particular, this implies $U^{(i)}_G=U^{(i)}_H\eqqcolon U^{(i)}$
    and $\mpc_G(u)=\mpc_H(u)\eqqcolon\mpc(u)$ for all $u\in U^{(i)}$.
  \item[(ii)] The partition $(V_i)_{i\in[\ell]}$ refines the partition
    $(V^{(i)}_u)_{u\in U^{(i)}}$.
  \item[(iii)] The partition $(W_i)_{i\in[\ell]}$ refines the partition
    $(W^{(i)}_u)_{u\in U^{(i)}}$.
  \item[(iv)] For every $u\in U^{(i)}$ and $j\in[\ell]$ we have
    $V_j\subseteq V_u^{(i)}\iff W_j\subseteq W_u^{(i)}$.
  \end{enumerate}
  Observe that (i) implies \eqref{eq:2}. 

  For the base step, we recall that $\mpc^{(1)}_G(v')=Z(G,v')\in\CB$,
  and that for every $c\in\CB_1$, we have
  \begin{align*}
    \{v'\in V(G)\mid Z(G,v')=c\}&=\bigcup_{i\in[\ell]\text{ with }c_i=c}V_i,\\
    \{w'\in V(H)\mid Z(H,w')=c\}&=\bigcup_{i\in[\ell]\text{ with }c_i=c}W_i.
  \end{align*}
  Thus both $D^{(1)}_G$ and $D^{(1)}_H$ consist of isolated nodes
  $u_1,\ldots,u_\ell$ with $\mpc_G(u_i)=\mpc_H(u_i)=c_i$, and we have
  $V^{(1)}_{u_i}= \bigcup_{i\in[\ell]\text{ with }c_i=c}V_i$ and
  $W^{(1)}_{u_i}= \bigcup_{i\in[\ell]\text{ with }c_i=c}W_i$. This
  implies assertions (i)--(iv) for $i=1$.

  For the inductive step $i\to i+1$, assume that we have proved
  (i)--(iv) for $i$. To prove it for $i+1$, we shall prove that for
  all $j\in[\ell]$  there is a color $c^{(i+1)}_j$ such that for all $v'\in V_j$ it
  holds that $\mpc_G^{(i+1)}(v')=c^{(i+1)}_j$ and for all $w'\in W_j$ it
  holds that $\mpc_H^{(i+1)}(w')=c^{(i+1)}_j$. Assertions  (i)--(iv)
  for $i+1$ follow.

  Let $j\in [\ell]$. We need to prove that for all $u\in U^{(i)}$
  there is a $a_{ju}\ge 0$ such that all $v'\in V_j$ have $a_{ju}$
  neighbours in $V_u^{(i)}$ and  all $w'\in W_j$ have $a_{ju}$
  neighbours in $W_u^{(i)}$. So let $u\in U^{(i)}$. By the induction
  hypothesis, there are $j_1,\ldots,j_k\in[\ell]$ such that
  $V_u^{(i)}=V_{j_1}\cup\ldots\cup V_{j_k}$ and
  $W_u^{(i)}=W_{j_1}\cup\ldots\cup W_{j_k}$. We let $a_{ju}\coloneqq
  A_{jj_1}+\ldots+A_{jj_k}$. As every $v'\in V_j$ has $A_{jj_p}$
  neighbours in
  $V_{j_p}$, it has $a_{ju}$ neighbours in $V_u^{(i)}$. Similarly,
  every $w'\in W_j$ has $a_{ju}$ neighbours in $W_u^{(i)}$.
\end{proof}

\begin{corollary}\label{cor:no1}
  Let $G,H\in \CG_{\CB}$ be graphs and $v\in V(G),w\in V(H)$. Then the following are equivalent:
  \begin{enumerate}
  \item $|G|=|H|$ and there is a weak node sketch $S$ such that both $(G,v)$ and
    $(H,w)$ realise $S$;
  \item $\mpc_G(v)=\mpc_H(w)$;
  \end{enumerate}
\end{corollary}

\begin{lemma}\label{lem:no2}
    There is a polynomial-time algorithm that, given the dag
    $D(\mpc_G(v))$ for some connected graph $G$ and  node $v\in V(G)$, 
    computes the weak node sketch $S_{\textup w}(G,v)$.
\end{lemma}

\begin{proof}
  The proof of the implication (3)$\implies$(2) of Lemma~\ref{lem:no1}
  describes how to construct $S_{\textup w}(G,v)$ from
  $D(\mpc^{(t)}_G(v))$ (without ever seeing $(G,v)$). It is easy to
  see that this construction can be turned into a polynomial time algorithm.
\end{proof}

The next lemma is a node-level variant of Lemma~\ref{lem:kiefer}.

\begin{lemma}\label{lem:nk}
  There is a polynomial-time algorithm that, given a weak node sketch
  $S$ and an $n\ge 1$ in unary, decides if there is a rooted graph $(G,v)$ of order $n$ realizing $S$
  and computes such a graph if there is.
\end{lemma}

\begin{proof}
  Let $S=(A,\boldsymbol c,k)$, where $A=(A_{ij})_{i,j\in[\ell]}\in\mathbb
  N^{\ell\times\ell}$, $\boldsymbol
  c=(c_1,\ldots,c_\ell)\in\mathcal B^\ell$, and $k\in[\ell]$. Observe that $S$ is not realisable if the graph $H\coloneqq([\ell],\{ij\mid A_{ij}>0\})$ is disconnected, because in any realisation $(G,v)$ of $S$ the matrix carries information about the connected component $G_v$. 
  So we first check if the graph $H$ is connected and reject if it is not.
  
  Observe next that $S=(A,\boldsymbol
  c,k)$ is realisable by a connected graph of order at most $n$ if and only if there is a tuple $\boldsymbol
  b\in\mathbb N_{>0}^\ell$ such that $\Sigma_{i=1}^\ell b_i\le n$ and the sketch $(A,\boldsymbol b,\boldsymbol c)$ is
  realisable. To find such a tuple $\boldsymbol b$, we need to satisfy $\Sigma_{i=1}^\ell b_i\le n$ and conditions (i) and
  (ii) of Lemma~\ref{lem:kiefer}, which amounts to solving a system of
  linear equations in the variables $b_i$. If there is no solution,
  then $S$ is not realisable by connected graph of order at most $n$. If we have found a solution $\boldsymbol b$,
  using Lemma~\ref{lem:kiefer} we compute a connected graph $G'$ and a partition
  $V_1,\ldots,V_\ell$ realising $(A,\boldsymbol b,\boldsymbol c)$. We pick an arbitrary node
  $v\in V_k$. Then $(G',v)$ realises $S$. However, we may have $|G'|<n$. If that is the case, we extend $G'$ by $n-|G'|$ isolated vertices to obtain a graph $G$. Then $(G,v)$ still realises $S$, and we have $|G|=n$.
\end{proof}

\begin{corollary}\label{cor:no}
  There is a polynomial-time algorithm that, given the dag
    $D(\mpc_G(v))$ for some graph $G$ and node $v\in V(G)$, computes a graph $G'$ of order $|G'|=|G|$ and node $v'\in V(G')$ such that $\mpc_{G'}(v')=\mpc_G(v)$.
  \end{corollary}

  \begin{proof}
    Let $D=D(\mpc_G(v))$ for some graph $G$ and node $v\in
    V(G)$, and let $n\coloneqq|G|$. Since $\mpc_G(v)=\mpc_G^{(2n)}(v)$, the dag $D$ has $2n+1$ levels, and hence we can compute $n$ from $D$. Using
    Lemma~\ref{lem:no2} we compute $S\coloneqq S_{\textup w}(G,v)$.
    
    Using
    Lemma~\ref{lem:nk}, we compute a rooted graph $(G',v')$ of order $|G'|=n$ realising
    $S_{\textup w}(G,v)$.  It follows from 
    Corollary~\ref{cor:no1} that $\mpc_{G'}(v')=\mpc_G(v)$.
  \end{proof}

\begin{proof}[Proof of Lemma~\ref{alem:crinv}]
  The implications (2)$\implies$(1) is trivial. We need to prove the converse

  Suppose that $F$ is mp-invariant. Then given $D(\mpc^{(t)}_G(v))$, using Corollary~\ref{cor:no}, we compute a graph $G'\in\CG_{\CB}$ and a node $v'\in V(G')$ such that $\mpc^{(t)}_{G'}(v')=\mpc^{(t)}_G(v)$ and $|G'|=|G|$. Then we compute $F(G',v')$. Since $F$ is CR-invariant, this is equal to the desired $F(G,v)$.

  Since the algorithm of Corollary~\ref{cor:no} runs in polynomial time, the additional statement on the running time follows.
\end{proof}

\subsection{A Version for Graph-Level Function}

In general, the message-passing color of a node does not determine the complete Color Refinement information of a graph. The following example shows that there are graphs $G$, $H$ of the same size and nodes $v\in V(G),w\in V(H)$ such that $\mpc^G(v)=\mpc_H(w)$ and yet Color Refinement distinguishes $G$ and $H$.

\begin{example}
    Let $G$ be a cycle of length $4$, and let $H$ be the disjoint union of a triangle and an isolated vertex. Then clearly Color Refinement distinguishes $G$ and $H$.
    Now let $v\in V(G)$ be arbitrary, and let $w$ be a vertex of the triangle of $H$. Then $\mpc^{(t)}_G(v)=\mpc^{(t)}_H(w)$ for all $t\in\Nat$.
\end{example}

However, this cannot happen for connected graphs.

\begin{lemma}\label{lem:nl2gl}
    Let $G,H$ be connected graphs of order $n\coloneqq|G|=|H|$, and let $v_0\in V(G),w_0\in V(H)$ such that $\mpc^{(t)}_G(v_0)=\mpc^{(t)}_H(w_0)$ for some $t\ge 2n$. Then Color Refinement does not distinguish $G$ and $H$.
\end{lemma}

\begin{proof}
It follows from Lemma~\ref{lem:no1} that $(G,v_0)$ and $(H,w_0)$ have the same weak node sketch, say,
\[
  (A,\boldsymbol c,k)\coloneqq S_{\textup w} (G,v_0)=S_{\textup w} (H,w_0),
\]
where $A\in \Nat^{\ell\times\ell}$ and $\boldsymbol c\in\CB^\ell$.

Suppose that the MPC-equivalence classes of $G$ are
$V_1,\ldots,V_\ell$ and that the MPC-equivalence classes of $H$ are
$W_1,\ldots,W_\ell$ (both in canonical order). For every $i\in[\ell]$, let $x_i\coloneqq |V_i|$ and $y_i\coloneqq|W_i|$. We need to prove that for all $i\in[\ell]$ we have $x_i=y_i$.

Recall that for all distinct $i,j\in[\ell]$, every $v\in V_i$ has exactly $A_{ij}$ neighbors in $V_j$. Hence the number of edges in $G$ between $V_i$ and $V_j$ is $x_iA_{ji}$. By the same reasoning applied from $j$ to $i$, the number of edges between $V_i$ and $V_j$ is also $x_jA_{ji}$. Thus
\begin{equation}
  \label{eq:1}
  x_iA_{ij}=x_jA_{ji}. 
\end{equation}
Similarly,
\begin{equation}
  \label{eq:3}
  y_iA_{ij}=y_jA_{ji}. 
\end{equation}

\medskip\noindent
\textit{Claim~1.}
For every $i\in[\ell]$ there is an $a_i\neq 0$ such that $x_i=a_i x_k$ and $y_i=a_iy_k$.

\medskip\noindent
\textit{Proof.} Let $i\in[\ell]$. If $i=k$, we let $a_i\coloneqq 1$, and the assertion is trivial. Otherwise, since $G$ is connected, there is an $m\ge 1$ and a seqeunce $i_1,\ldots,i_m$ such that $i_1=i$ and $i_m=k$ and $A_{i_ji_{j+1}}\neq 0$ for $1\le j<m$. By \eqref{eq:1}, we have $x_j\coloneqq\frac{A_{i_{j+1}i_j}}{A_{i_ji_{j+1}}}x_{j+1}$. Thus with
\[
  a_i\coloneqq \prod_{j=1}^{m-1}\frac{A_{i_{j+1}i_j}}{A_{i_ji_{j+1}}}
\]
we have $x_i=a_ix_k$. Similarly, using \eqref{eq:3}, we obtain $y_i=a_ix_k$.
This proves the claim.

\medskip
Since $\Sigma_{i=1}^kx_i=|G|=|H|=\Sigma_{i=1}^ky_i$, by the claim we thus have
\[
  x_k\Sigma_{i=1}^ka_i=y_k\Sigma_{i=1}^ka_i,
\]
which implies $x_k=y_k$ and, again by the claim, $x_i=a_ix_k=a_iy_k=y_i$ for all $i$.
\end{proof}

\begin{corollary}\label{cor:nl2gl}
    There is a polynomial-time algorithm that, given the dag
    $D(\mpc_G(v))$ for some connected graph $G$ and node $v\in V(G)$, computes a graph $G'$ such that Color Refinement does not distinguish $G$ and $G'$.
\end{corollary}

\subsection*{The Weisfeiler-Leman Algorithm}
Following \cite{Grohe21}, we distinguish between the Color Refinement algorithm and the Weis\-feiler-Leman algorithm.
In the literature, this distinction is usually not made, and what we call Color Refinement here is called Weisfeiler-Leman. Regardless of the terminology, there are two different algorithms, and the difference between them is often overlooked or ignored, because in many situation it is not relevant. However, it does make a difference here. (We refer the reader to \cite{Grohe21} for a discussion and an example illustrating the difference.)

Where Color Refinement collects the local message-passing information
along the edges of a graph,
Weisfeiler-Leman no longer restrict the information flow to edges, but considers the information flow along non-edges as well.
For every $t\ge 0$ and $v\in V(G)$, we define a color $\wl_G^{(t)}(v)$
inductively as follows:
  \begin{itemize}
  \item $\wl^{(0)}_G(v)\coloneqq \mpc_G^{(0)}=Z(G,v).
    $
  \item $
      \wl^{(t+1)}_G(v)\coloneqq\big(\wl^{(t)}_G(v),\lmulti
      \wl^{(t)}_G(w)\mid w\in N_G(v)\rmulti, \lmulti
      \wl^{(t)}_G(w)\mid w\in V(G)\setminus N_G(v)\rmulti\big).
    $
  \end{itemize}
For WL, we define the final color to be
$\wl_G(v)\coloneqq\wl_G^{(|G|)}(v)$. We define set $\WL^{(t)}_{n,k}$ analogous to the corresponding set $\MPC^{(t)}_{n,k}$.
We can also represent the WL colors by a dag, the simplest way of
doing this is to also introduce edges with negative labels to
represent the multiset of colors of non-neighbours.

Although it seems that the non-local message passing in WL adds
considerable power, this is actually not the case. In particular, CR
and WL have exactly the same power when it comes to distinguishing
graphs: it can be shown that for all $G,H$, CR distinguishes $G,H$ if and only if WL
distinguishes $G,H$ (see \cite{Grohe21} for a proof).

Note, however, that this is no longer the case on the node level. For
example, if $G$ is cycle of length $3$ and $H$ a cycle of length $4$
then for all $v\in V(G)$ and $w\in V(H)$ it holds that
$\mpc_G(v)=\mpc_H(w)$ and $\wl_G(v)\neq\wl_H(w)$. The distinguishing power of
WL on the node level exactly corresponds to that of GNNs with global readout.

There is also a version of Lemma~\ref{lem:crinv} for the WL-algorithm. A
feature transformation $F: \CG_{\CB}\rightarrow \CZ_{\CB}$ is
\emph{WL-invariant} if for all graphs $G,H\in \CG_{\CB}$ and nodes
$v\in V(G), w\in V(H)$, if $\wl^{(t)}_G(v)=\wl^{(t)}_H(w)$ for all
$t\ge 1$ then $F(G,v)=F(H,w)$. Note that each mp-invariant feature
transformation is also WL-invariant, but that the converse does not
hold.

\begin{lemma}\label{lem:wlinv}
  Let $F: \CG_{\CB}\rightarrow \CZ_{\CB}$ be computable.
  Then the following are equivalent.
  \begin{enumerate}
  \item $F$ is WL-invariant.
  \item There is an algorithm that computes $F(G,v)$ from $\wl_G(v)$. 
  \item There is an algorithm that computes $F(G,v)$ from $\wl^{(t)}_G(v)$ for an arbitrary $t\ge|G|$. 
\end{enumerate}
  Furthermore, if $F$ is computable in time $T(n)$ then algorithms
  in (2) and (3) can be constructed to run in time $T(n)+\poly(n)$,
  and conversely, if the algorithm in either (2) or (3) runs in time
  $T(n)$ then $F$ is computable in time $T(n)+\poly(n)$.
\end{lemma}

The proof of this lemma is similar to the proof of
Lemma~\ref{alem:crinv}. Instead of the weak node sketch, we work with
the \emph{node sketch} $S(G,v)=(A,\boldsymbol b,\boldsymbol c,k)$, where $(A,
\boldsymbol b,\boldsymbol c)=S(G)$ is the sketch of $G$ and $k$ the
index of the class of the coarsest stable partition that contains $v$.

\section{Main Result}
For a binary string $\CB_k\ni x=b_1,\ldots,b_k$ define its integer value $\BTI(x)\coloneqq\Sigma_{i\in[k]}b_i2^{i-1}$, and for binary strings $x_1,x_2\in\CB$ define their binary sum $x_1+x_2\coloneqq \BTI^{-1}(\BTI(x_1)+\BTI(x_2))$. When clear from the context, we may refer to $x\in\Nat$ while meaning its binary representation $\BTI^{-1}(x)$.

\begin{definition}
    A $\relu$-activated Multilayer Perceptron (MLP) $F=(l_1,\ldots,l_m), l_i=(w_i,b_i)$, of I/O dimensions $d_{in};d_{out}$, and depth $m$, is a sequence of rational matrices $w_i$ and bias vectors $b_i$ such that 
$$\dim(w_1)(2)=d_{in}, \dim(w_m)(1) = d_{out},$$$$\;\forall i>1 \; \dim(w_i)(2)=\dim(w_{i-1})(1),$$
$$ \forall i\in[m] \dim(b_i)=\dim(w_i)(1)$$
It defines a function $f_F(x)\coloneqq$$$\relu(w_m(...\relu(w_2(\relu(w_1(x)+b_1))+b_2)...)+b_m)$$
When clear from the context, we may use $F(x)$ to denote $f_F(x)$.
\end{definition}

\begin{figure*}[t!]
\centering
  \includegraphics[trim={2.5cm 5.5cm 2.5cm 4.5cm},clip=true,width=.99\linewidth]{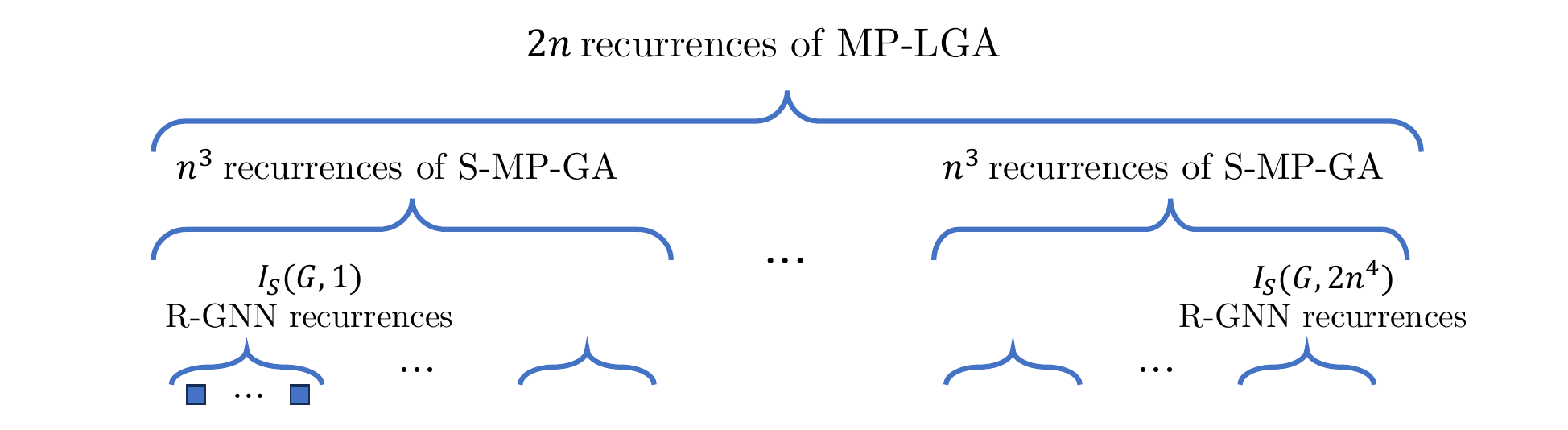}
  \caption{Let $G\in\CG_{\CB_k}$ and define $n\coloneqq |G|$. An illustration for reducing the construction of $\mpc_G(v)$ to R-GNN. First it is reduced to $O(n)$ iterations of MP-LGA.
  Then, each iteration of MP-LGA is reduced to $O(n^3)$ iterations of S-MP-GA. Let that S-MP-GA be $S$, and let $T_S(G,i)$ be the number of Turing machine steps required to compute the $i^{th}$ iteration of $S$ when operating on $G$. Then, iteration $i,i\in[2n^4]$ is reduced to $O(I_S(G,i))$ recurrences of an R-GNN, where $I_S(G,i)\coloneqq T_S(G,i)+(n^3\log(n)+kn)^2$. The $(n^3\log(n)+kn)^2$ overhead is for translating the sum of messages from RB to RQ.}
  \label{fig:reduction_stages}
\end{figure*}

\begin{theorem}\label{thm:sizemust}
  There exists a feature transformation $F: \CG_{\CB}\rightarrow \CZ_{\CB}$ such that for every message-passing algorithm $\CA$ where the graph-order input is omitted, there exist $G\in\CG_{\CB}, v\in V(G)$ for which $\CA(G,v)\neq F(G,v)$.
\end{theorem}
\begin{proof}
    Consider the function $F$ defined by $F(G,v)=1$ if $v$ is contained in a cycle of $G$ and $F(G,v)=0$ otherwise. It is not hard to see that $F$ is mp-invariant. 

    Suppose that there is a message-passing algorithm $\CA$, where the graph-order input is omitted, that computes $F$.
    Consider the computation of $\CA$ on a cycle. Regardless of the length of the cycle, the computation will be the same, because for all cycle $C,C'$, all nodes $v\in V(C)$, $v'\in V(C')$, and all $t\in\mathbb N$ it holds that $\mpc_C^{(t)}(v)=\mpc_{C'}^{(t)}(v')$. Hence there is an $I\in\Nat$ such that for all cycles $C$ and nodes $v\in V(C)$ we have $I_v=I$, where $I_v\in\Nat$ is the finishing iteration of $\CA$ on $C,v$. That is, the computation terminates after $I$ rounds and returns the value $F(C,v)=1$.

    Now consider a long path $P$ of even length $\ge 2I$ and let $w$ be the middle node of this path. As the neighborhood of radius $I$ of $w$ is identical to the neighborhood to a node $v$ on a cycle $C$ of length $\ge 2I$, we have $\CA^{(i)}(P,w)=\CA^{(i)}(C,v)$ for all $i\le I$. Hence $I_w=I$ and $\CA(P,w)=\CA(C,v)=1$, hence $\CA(P,w)\neq F(P,w)$.
\end{proof}

The next three lemmas state the reductions to intermediate models. \Cref{fig:reduction_stages} illustrates the composition of these reductions from a recurrences perspective.
\lemmpcgctompgc*
\begin{proof}
    The idea is to construct $\mpc_G(v)$ step by step in the first $2|G|$ applications of $f$. Then, in the $(2|G|+1)$ application, compute the function determined by $M$, on the constructed $\mpc_G(v)$. The state of the computation remembers the last iteration-number $t$ (up to $2|G|+1$), the graph size, and $\mpc_G^{(t)}(v)$. The required sum-of-messages length limit is implied by the dag-construction description in \Cref{subsec:dagconstruct}.
    For an encoding of a triplet of binary strings $x=\te(b_1,b_2,b_3),b_i\in\CB$, define:    
    \\\phantom{} $\quad\quad\quad\quad x.t\coloneqq \BTI(b_1)$, representing the iteration-number part
    \\\phantom{} $\quad\quad\quad\quad x.s\coloneqq \BTI(b_2)$, representing the graph size part
    \\\phantom{} $\quad\quad\quad\quad x.d\coloneqq b_3$, representing the $\de(\mpc_G^{(x.t)}(v))$ part.    
    \\Let $\de(c_1),\ldots,\de(c_l), \forall i\in[l]\; c_i\in \MPC^{(t)}_{|G|,k}\;$ be the dag encodings of a vertex $v$ and its neighbors colors after $t$ Color Refinement iterations, for some $k,t$. We define the operation of combining these encodings into the dag encoding of the next-iteration color of $v$.
    \begin{center}
        $\dc(\de(c_1),\lmulti\de(c_2),\ldots,\de(c_l)\rmulti)\coloneqq \de(c_1,\lmulti(c_2),\ldots,(c_l)\rmulti)$
    \end{center}
    We define $(A,f)$ as follows:
    $A=0$, representing an initial iteration-number of zero. $f(x_1,x_2)\coloneqq$
    $$
        \begin{cases}
            (\te(x_1.t + 1,x_1.s,\de(x_1.d,x_2)),\dc(x_1.d,x_2))&\\ &x_1.t\leq x_1.s\\
            (\te(x_1.t + 1,x_1.s,x_1.d),M(x_1.d))&\\ & x_1.t=x_1.s+1
        \end{cases}
    $$            
\end{proof}

\lemmpgctompgc*
\begin{proof}
    Unlike in the proof of Lemma \ref{lem:mpcgc_to_mpgc}, the emulating function $f'$ is not very concise, hence we define it in pseudo-code style, in \Cref{lst:mpgc_to_mpgc_one}.
    Each recurrence of $C$, where a node simply receives the multiset of its neighbors' features, requires $O(|G|^3)$ recurrences of $C'$, where a node receives the sum of whatever its neighbors are sending, in order to extract the neighbors' individual features. The idea of such a phase of $O(|G|^3)$ recurrences is as follows:
    \begin{itemize}
    \setlength\itemsep{0.1em}
        \item [1.] For the first $|G|^2$ recurrences, vertices propagate max(own value, neighbors' messages average). The neighbors' average can be computed by dividing the sum of values by the sum of '1' each sender sends a dedicated dimension. Note that if there are two vertices with different values, the lower-value one will necessarily perceive an average higher than its own value - even if they are farthest from each other, after at most $|G|$ recurrences. Whenever a vertex perceives a higher average than its own value it temporarily disables itself and propagates the average, until the end of the first $|G|^2$ recurrences. This is implemented in the 'find\_max' code. Hence, after $|G|^2$ recurrences, it is guaranteed that the vertices left enabled are exactly those with the maximum value in the whole graph - excluding those already counted for and permanently disabled (see below). The $|G|^2 + 1$ recurrence is used so that each vertex operates according to whether it is one of the max-value vertices. This is implemented in the 'send\_if\_max' code. Then, in the $|G|^2 + 2$ recurrence each vertex knows that the average it receives is actually the maximum value, and the count is the number of its neighbors with that value, and adds that information to the constructed multiset of neighbors' values. This is implemented in the 'receive\_max' code. Then, all max-value vertices permanently disable themselves, all temporarily-disabled vertices re-enable themselves, and another $|G|^2$ recurrences phase starts - to reveal the next maximum-value and counts.
        \item [2.] Since there are at most $|G|$ distinct values, after $|G|$ iterations of the process above, every vertex has finished constructing the multiset of its neighbors values. All is left to do then is to apply $f$ of the MP-LGA on the (current value and) constructed multiset, and update the vertex's value to be the output of $f$.
    \end{itemize}
    The required message length limit is implied by the definition of the algorithm and the dag-construction description in \Cref{subsec:dagconstruct}.
\end{proof}

\begin{figure*}[t!]
\centering
  \includegraphics[trim={1cm 4cm 1cm 4cm},clip=true,width=.99\linewidth]{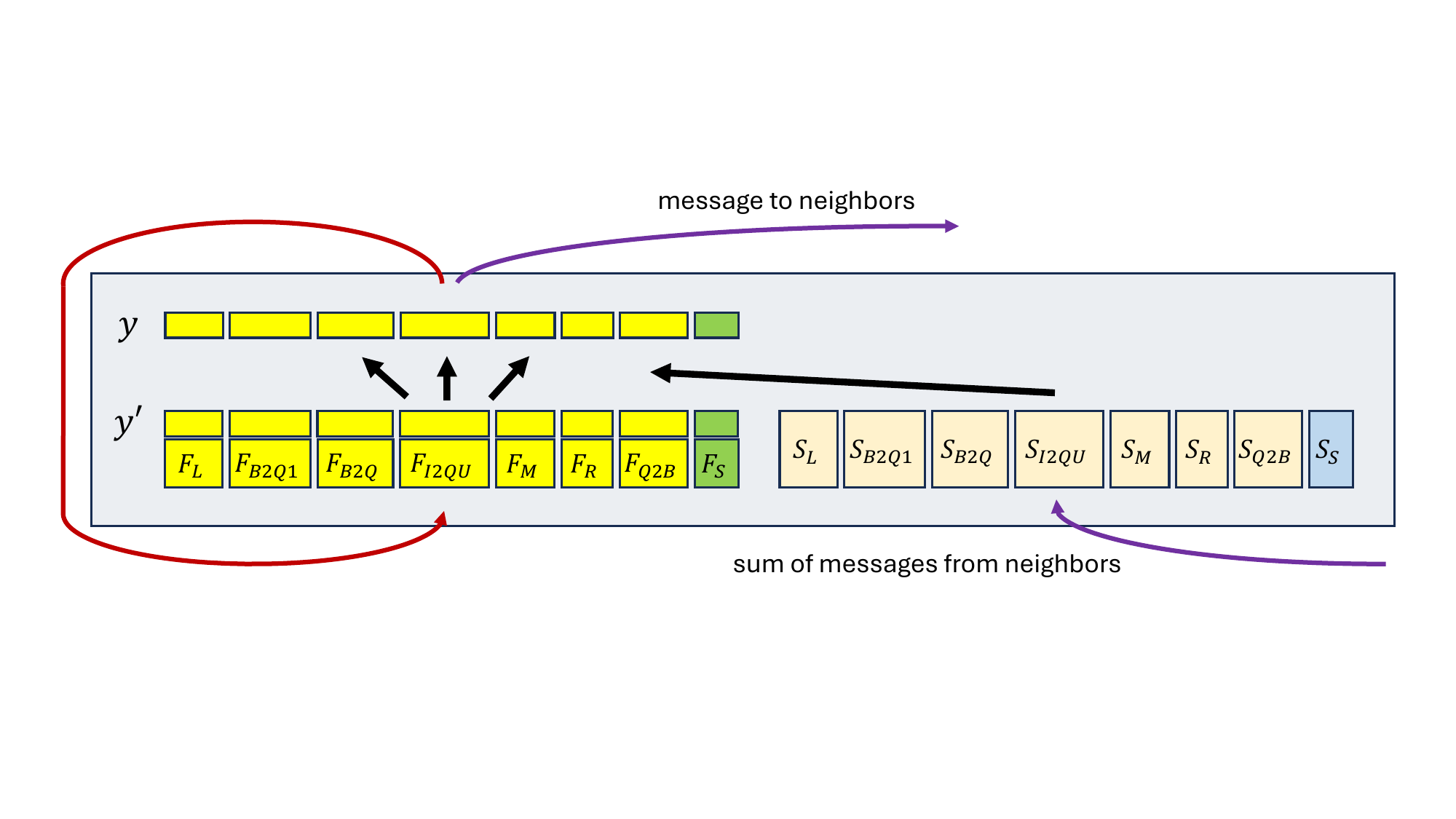}
  \caption{The structure of $F$. In yellow are all the switched sub-networks that pass control and data from one to another. In green is the synchronization sub-network that runs continuously. In beige and blue are the dimensions that assume the sum of neighbors' sub-networks output in the previous recurrence. Layer $y'$ is the outputs of the individual sub-networks. These, together with the neighbors' sum, are inter-routed using layers $[y'+1..y]$, to create the desired interoperability. Finally, the output of layer $y$ is sent to neighbors and also becomes (half) the input for the next recurrence.}
  \label{fig:construction}
\end{figure*}

\lemmsmpgctorgnn*
\begin{proof}
Note that an R-GNN is essentially an extension of a recurrent MLP to the sum-message-passing setting. In \cite{siegelmann1992computational} it is shown that recurrent MLPs are Turing-complete. Assume $C=(A,f)$ and let $M$ be the Turing-machine that computes $f$, we would like to use the result in \cite{siegelmann1992computational} and emulate $M$ using the recurrent MLP in an R-GNN. However, this requires overcoming two significant gaps:
\begin{itemize}
    \setlength\itemsep{0.1em}
\item[1.] An encoding gap. In \cite{siegelmann1992computational} the emulation of a Turing machine is done by emulating a two-stack machine where a stack's content is always represented as a value in $\RQ$. Since $\RQ$ is not closed under summation, a naive attempt to use the sum of the neighbors' stacks directly - as input to a Turing machine emulation is doomed to fail: The sum may be an invalid input and consequently the output will be wrong. To overcome this, we precede and proceed the Turing-machine-emulation recurrent MLP with recurrent sub-networks that compute translations
$$\RB2\RQ:\RB\rightarrow\RQ,\; \RB2\RQ\coloneqq\rqe\circ\rbe^{-1}$$$$\RQ2\RB:\RQ\rightarrow\RB,\; \RQ2\RB\coloneqq \rqe\circ\rbe^{-1}$$
The former translates the received sum of messages, and the latter translates the new message to be sent.
The depth of these networks is constant and the number of recurrences required for each translation is quadratic for $\RB2\RQ$ and linear for $\RQ2\RB$, in the number of translated bits. Existence of such recurrent MLPs is not trivial - the result in \cite{siegelmann1992computational} assumes an input in $\RQ$ for a reason.

The messages that are used in our algorithm have a fixed-length, and have two parts which the receiver should be able to read separately. When multiple messages are summed, those two parts can potentially interfere. However, we make sure that the message length is large enough to contain the sums of both parts separately, and assume that they are written in two separate parts of the message, hence it is guaranteed that there will be no interference. 
Define $L_{\de}(n,k)\coloneqq$ $$\max\big(|\de(\mpc_G(v))| : G\in\CG_{\CB_k}, |G|=n, v\in V(G)\big)$$
the maximum length of the dag encoding of a vertex in a graph of size $n$ with initial features of length $k$. The fixed message length is $3kn^4$, which is more than the maximum-possibly-required $\log(n(2^{L_{\de}(n,k)}+2^1))$ - the length of the sum of $n$ complete 
$\mpc$-plus-a-$0/1$-indicator. The sub-networks are designed to translate those messages. To do that, one of the inputs of the $\RB2\RQ$ sub-network is the fixed message-length. Note that $k$ is known before constructing the overall network, and $n$ is given as an initial feature of the vertex, however we provide a stronger result by describing a network that receives also $k$ as an input hence supports all initial-feature lengths. We construct a dedicated recurrent sub-network that computes $3kn^4$ at the very beginning of the overall computation, which can then be used by the sub-networks that require this value as one of their inputs.

In addition to the translation of the incoming and outgoing messages, it is required at the beginning of the run to translate the message-length; graph-size; and initial-feature - from an integer and a value in $\RB$ to one value in $\RQ$, to be used by the Turing-machine sub-network. This is done in two stages: First, another copy of an $\RB2\RQ$ translator translates the feature, and then a new kind of sub-network does both the translation of the graph size and the combination of the two $\RQ$ values - in a way that they can be separated by the Turing machine emulation.

Finally, As the two final-output dimensions of an R-GNN - a "finished" indicator and the final feature of the vertex - are defined in \Cref{def:spiral_gnn} to be a $0/1$ value and a value in $\RB$, we construct another translating network that translates those final outputs from $\RQ$, when the whole computation of the R-GNN is about to finish.

\item[2.]A synchronization gap. In an S-MP-GA computation, nodes are synchronized by definition: A node's $i^\th$ received sum-of-messages is the sum of its neighbors $i^\th$ sent message. On the other hand, in an emulating R-GNN that is based on \cite{siegelmann1992computational}, every Turing machine step corresponds to a recurrence and - being a GNN - every recurrence entails a message pass, that is, a potential Send/Receive. A naive attempt of a vertex to start a new computation whenever its emulated Turing machine has finished, hence considering the next message-pass as the sum of its neighbors' current-computations' results, will not work: The (emulated) Turing machines of different nodes with different inputs may require a different number of recurrences to complete their computations.
To overcome this, we augment the recurrent MLP described thus far with a recurrent MLP that synchronizes the start of a new computation,  across all nodes. That MLP runs for a the same number of recurrences for all nodes - regardless of differences in their inputs, hence, in itself it does not require synchronization.
\end{itemize}
Overall, the recurrent MLP $F$ consists of $8$ recurrent sub-networks:
\\$F_L,F_{B2Q_1},F_{I2QU},F_{B2Q},F_M,F_R,F_{Q2B},F_S$. For a graph and vertex $G\in\CG_{\CB_k},\ v\in V(G)$, the  computation flow is as follows:
\begin{itemize}
    \setlength\itemsep{0.01em}
    \item [a.] $F_L$ computes the length of the messages, $3k|G|^4$, to be used by $F_{B2Q}$, then passes control to (b).
    \item [b.] $F_{B2Q_1}$ translates $Z(G,v)$ from $\RB$ to $\RQ$, to be used by $F_{I2QU}$, then passes control to (c).
    \item [c.] $F_{I2QU}$ translates $|G|$ to its unary representation in $\RQ$ - base-4 '3' digits, and concatenates it to the base-4 '1' digit followed by the result of (b). That is, $x\mapsto (\Sigma_{i\in[x]}\frac{3}{4^i}) + \frac{1}{4^{x+1}} + \frac{1}{4^{x+2}}\rqe(\rbe^{-1}\big(Z(G,v))\big)$. The output is to be used by $F_M$. Then it passes control to (1.1).
    \item [1.1.] $F_{B2Q}$ translates the sum of messages, from $\RB$ to $\RQ$, then passes control to (1.2).
    \item [1.2.] $F_M$ computes the Turing machine function, then passes control to (1.3).
    \item [1.3.] $F_{R}$ Checks whether the overall computation - the function computed by the R-GNN - is finished, and if it is, translates the final output - final feature of the vertex - from $\RQ$ to $\RB$. Note that in such case this output is "locked" - it will not change over subsequent recurrences. If the overall computation is not finished, control is passed to (1.4).
    \item [1.4.] $F_{Q2B}$ translates the new message to send, computed by $F_M$, from $\RQ$ to $\RB$, then stops.
    \item [2.] In parallel to (1), $F_{S}$ operates continuously. Every $2n$ recurrences it produces a signal whether all nodes' previous computation is finished already $n$ recurrences before. If the signal is positive, the sequence of (1) is restarted.
\end{itemize}

We proceed to formalize the structure of the whole R-GNN network, that is, the structure of each sub-network and how theses sub-networks are put together. We describe them not necessarily in their order of operation - described above - but also in order of similarity in functionality.

We would like the sub-networks $F_L,F_{B2Q_1},F_{I2QU},F_{B2Q},F_M,F_R,F_{Q2B}$ to operate in two modes, 'On' - where they compute what they are supposed to (probably over multiple recurrences) and have the results as the values of certain output neurons, and 'Off' - where the results-outputs are remained unchanged. Switching to 'On' mode should be initiated by an external change to an 'On/Off' switch neuron, at which time certain preconditions must hold for the rest of the state neurons of the sub-network in order to assure that it indeed computes what it is supposed to. Switching to 'Off' mode is part of the sub-network's results when finishing its current computation, which in turn causes another sub-network to switch on, and so on and so forth. The goal is that at any recurrence of the whole R-GNN, only one sub-network, excluding $F_S$, is actually computing its function. We formalize the concept of a switched network, in the following definition.
\begin{definition}\label{def:switch_net}
    A \emph{switched} recurrent MLP (switched MLP) $F=(((w_1,b_1),\ldots,(w_m,b_m)),D,f)$, 
    of $d$ dimension, is a sequence of rational matrices and bias vectors - similar to a recurrent MLP; A set 
    $D\subseteq \{0,1\}\times\{0,1\}\times\Rat^{d_I+d_O}\times\Rat^{d-d_O-d_I-2}$;\\ and a function $f:\Rat^{d_I}\rightarrow\Rat^{d_O}, 2+d_I+d_O\leq d$. The $d-d_I-d_O$ dimensions are neurons that only remember the state of the computation. The semantics are that the first input dimension is a switch, the second is an indication if the switch has turned off from previous iteration, $D$ is a set of valid initial states, and $f$ is a target function whose input(s) start at the third input dimension, and whose output(s) start after the inputs. More formally:
    \begin{itemize}
    \setlength\itemsep{0.1em}
        \item [1.] $\forall x\in D \;\; x(1)=0\Rightarrow $
        \\$(f^{(1)}_{F}(x)(1)=0\wedge f^{(1)}_{F}(x)(2)=0\wedge f^{(1)}_{F}(x)[2+d_I+1,2+d_I+d_O]=x[2+d_I+1,2+d_I+d_O] \wedge f^{(1)}_{F}(x)\in D)$
        \\That is, for every valid input such that the first dimension is $0$, the MLP remains "switched off": Switch is 'Off', turned-off indicator is $0$, target-function outputs maintain their value, and the overall output is a valid initial state - ready for being switched on.
        \item [2.] For $x\in D$, define $t_x\coloneqq min\big(t : \forall t'\geq t\; f^{(t')}_{F}(x)=f^{(t)}_{F}(x)\big) - 1$.
        Then, $$\forall x\in D\;\; x(1)=1\Rightarrow$$ $$(f^{(t_x)}_{F}(x)(1)=0\ \wedge\ f^{(t_x)}_{F}(x)(2)=$$$$1\ \wedge\ f^{(t_x)}_{F}(x)[2+d_I+1,2+d_I+d_O]=$$$$f(x[3,2+d_I])\ \wedge\ f^{(t_x)}_{F}(x)\in D)$$
        \\That is, for every valid input such that the first dimension is $1$, the MLP is "switched on" and: After enough recurrences, the first output turns $0$ - switching off; the switched-off indicator turns on; the function outputs are the value of the target function applied to the function inputs; and the overall output is a valid initial state. Note that by (1) we have that $f^{(t_x+1)}(x)$ is the same as $f^{(t_x)}(x)$ except for the second dimension which is $0$ and not $1$, because the switch does not turn off in recurrence $t_x+1$ - it was already off.
    \end{itemize}
\end{definition}
Note that while an external input is required to initiate a computation - set on the 'On/Off' switch and initialize the inputs, it is only a preceding stage to the network's operation and not in the scope of its definition. This is in contrast to \Cref{def:external_in_net} where new external inputs are used in every recurrence throughout the computation and hence are part of the networks' specification.

\subsubsection*{Constructing $F_M$}
Remember that we wish the R-GNN to compute the same function as a given S-MP-GA $C=(B,h)$. The first sub-network we describe is the main switched sub-network: At the core, it is a recurrent MLP that emulates the Turing-machine that computes $h$, which exists by \cite{siegelmann1992computational}. Then, that MLP is wrapped so to make it a switched recurrent MLP.
\begin{lemma}
    Let $f:\CB^4\rightarrow \CB^4$ be a computable function, and let $g:\RQ^4\rightarrow \RQ^4$ such that 
    $$\forall x\in\CB^4\;\;g\big(\rqe(x)\big)=\big(\rqe(f(x))\big)$$
    Let $S\coloneqq \{y\in\RQ^4 \mid \exists x\in\CB^4\; y=\rqe(x)\}$ be the set of valid inputs to $g$.
    Define $$\forall d'\in\Nat\;\;D_{d'}\coloneqq \{x \mid x\in \{0,1\}^2\times S\times\RQ^4\times\{0,1\}\times S\times\{0\}^{d'-4}\}$$
    Then, for some $d'\in\Nat$ there exist a $(d'+11)$-dimension switched network 
    \begin{center}$F=\Big(\big((w_1,b_1),\ldots,(w_m,b_m)\big),D_{d'},g\Big)$\end{center}
\end{lemma}
\begin{proof}
    Let $d'\in\Nat$ so there exists a $d'$-dimension recurrent MLP $F'=(l'_1,\ldots,l'_{m'})$ such that 
    $$\forall x\in S\times\{0\}^{d'-4}\;\exists k\in\Nat :$$$$ (\forall t<k\; F'^{(t)}(x)(5)=0),\; F'^{(k)}(x)(5)=1,\; F'^{(k)}(x)[1,4]=g(x)$$
    Such $F'$ exists by \cite[Thm 2 and Section 4.3]{siegelmann1992computational}.
    We define a switched network $$F=\big((l_1,\ldots,l_{m}),D_{d'},g\big)$$ of dimension $d\coloneqq (d'+11)$ and depth $m\coloneqq (m'+5)$, as follows. The idea is to wrap a copy of $F'$ with 2 pre-processing and 3 post-processing layers, and additional dimensions - that relay the information between the pre and post processing parts. Note that later we describe how $F_M$ itself is wrapped when combined into the whole network of the R-GNN.
    \\We denote: The $d$-dimension input vector to the first layer by $x$; the $d$-dimension input vector to the third layer - the layer in which the copy of $F'$ starts, by $x'$; the output vector of layer $m'+2$, in which the copy of $F'$  ends, by $y'$; and the $d$-dimension output of the last layer, which is the input vector for the next recurrence, by $y$.
    
    The first $11$ dimensions represent $F$'s switch; switch-turned-off indicator; function inputs; function outputs; and previous-recurrence switch-value. Initially, $x(1)=x(2)=x(11)=0$ i.e. switch-related values are $0$. When $F$ is swithced on i.e. $x(1)=1,x(2)=0,x(11)=0$, we want the inputs to become $F'$ inputs and we want to reset the rest of $F'$ dimensions to $0$, and so we define
    $$x'[12,15]\coloneqq \lsig(x[3,6]-\lsig(1-\relu(x(1)-x(11)))) + $$$$\lsig(x[12,15]-\lsig(\relu(x(1)-x(11))))$$
    $$x'[16,d]\coloneqq \lsig(x[16,d]-\lsig(\relu(x(1)-x(11))))$$
    We also want the switch-turned-off indicator to always remember the switch state at the beginning of the recurrence, and so we define $x'(2)\coloneqq x(1)$.        
    Considering that the activation function of our network is $\relu$, $x'$ can be computed from $x$ using at most 2 layers. These would be $l_1,l_2$.
    
    On the output end, if the input switch was $0$, we want all outputs of $F'$ to be zero - so it is ready for a new run. Otherwise: If $F'$ is finished i.e. $y'(16)=1$ then we want $F'$ function outputs i.e. $y'[12,15]$ to be the output of $F$ i.e. $y[7,10]$, its switch to turn off, and the switch-turned-off indicator to be $1$. If $F'$ is not finished then we want to pass all outputs of $F'$ as they are to next recurrence. The above logic can be implemented as follows:
    \begin{align*}
        &y(1)=\relu(y'(1)-y'(16)) \text{ (if $F'$ finished then turn off)}\\
        &y(2)=\relu(y'(2)-\relu(y'(1)-y'(16))) \text{ (if $F'$ just turned off)}\\
        &y(11)=\relu(y'(1)-y'(16))\\
        &y[7,10]= \relu\Big(y'[12,15]-\relu\big(1-y'(16))\Big)+ \\
        &\relu\Big(y'[7,10]-y'(16)\Big)
    \end{align*}
    
    Considering that the activation function of our network is $\relu$, $y$ can be computed from $y'$ using at most 3 layers. These would be $w_{2+m'+1},w_{2+m'+2},w_{2+m'+3}$.        
    Layers $[3..m'+2]$ are essentially $2$ separate stack-of-layers side-by-side, as follows:    
    \begin{itemize}
    \setlength\itemsep{0.1em}
        \item [1.] The first stack is $F$'s (switch, switch-turned-off indicator, function inputs, function outputs, previous switch), in dimensions $[1..11]$, each simply passing on from layer to layer.
        \item [2.] The second column is $F'$ as is, set in layers $[3..m'+2]$.
    \end{itemize}    
\end{proof}

For $F_M$, we did not explicitly describe the core network that implements the required functionality. Instead, we deduced its existence, from the result in \cite{siegelmann1992computational}. However, the mode of operation of a network a-la \cite{siegelmann1992computational} does not match the mode of operation of an R-GNN. As the combined functionality of $F_L,F_{B2Q_1},F_{I2QU},F_{B2Q},F_M,F_R,F_{Q2B}$ is meant to bridge that gap, the descriptions of these sub-networks cannot themselves be based on \cite{siegelmann1992computational} since this would just move the gap to a different location in the algorithm. Hence, we describe each of these sub-networks explicitly: We provide a pseudo-code that implements the required functionality when ran recurrently, and meets a certain specification which implies implementabilty by an MLP. The specification consists of a structure and an instruction-set, which together define an MLP implementation, as follows.
\subsubsection*{Programming MLPs}
\begin{definition}\label{def:MLP_code}
An \emph{MLP Code} having $d$ state-variables i.e. variables that should maintain their value from the end of one recurrence to the beginning of the next, has the following properties which together imply a $d$-dimension MLP implementation.
\begin{itemize}
    \setlength\itemsep{0.1em}
    \item [-] The code has two sections:
    \\An 'Initialize' section. There, the $d$ state-variables, which correspond to the I/O neurons of the MLP, are defined. Their names are prefixed by 's\_', and their initial values - before the R-GNN run starts - are defined as well. These must be non-negative.
    \\A 'RecurrentOp' section. There, is the code that defines a single recurrence of an implementing MLP. It consists solely of statements that conform to one of the following definitions of an \emph{expression}. The invariant of these definitions is that for every expression there exists an MLP sub-structure whose inputs are the statement's operands and whose output is the value of the statement.
    \begin{itemize}
    \setlength\itemsep{0.1em}
    \item [-] The state-variables are expressions. They correspond to the input neurons, and to subsequent dedicated neurons in each layer. That is, for each state-variable there is a corresponding neuron in each layer.
    \item [-] For the sake of code-readability, intermediate variables may also be used, and are considered expressions. Declared using the keyword 'var', they correspond to a specific neuron whose value is defined by an assignment to the variable.
    \item [-] Let $x_1,\ldots,x_n$ be expressions, let $w_1\ldots,w_n\in\Rat,b_1\ldots,b_n\in\Rat$, and define $x\coloneqq\Sigma_{i\in[n]}w_ix_i+b_i$, then:
    \begin{itemize}
        \item [1.] Both $\relu(x),\lsig(x)$ are expressions, regardless of the sign of $x$. Note that $\lsig$ is expressible by combining two $\relu$s.
        \item [2.] If $x\geq0$ then $x$ is an expression.
    \end{itemize}    
    \item [-] Let $v,x$ be a variable (expression) and an expression, then $v=x$ is an expression. Note that such assignment is implemented by feeding the neuron that represents $x$ into a next-layer neuron that represents $v$.
    \item [-] Let $v$ be a variable, and let $0\leq x\leq 1, s\in\{0,1\}$ be expressions representing a change and a state. Then, the conditional addition $$v.\text{increase\_if}(s,x)\coloneqq 
    \begin{cases}
        v+x & s=1\\
        v   & s=0
    \end{cases}
    $$
    is an expression, and if $v-x>0$ then also the conditional subtraction, 
    $$v.\text{decrease\_if}(s,x)\coloneqq 
    \begin{cases}
        v-x & s=1\\
        v   & s=0
    \end{cases}
    $$
    is an expression.
    Note that due to the restrictions $0\leq x\leq 1, s\in\{0,1\}$, $v.\text{increase\_if}(s,x)\equiv v + \relu(x-(1-s))$, Similar argument holds for $v.\text{decrease\_if}(s,x)$.
    \item [-] Let $0\leq v\leq 1$ be a variable, let $a\in\Nat$ be a constant, and let $s\in\{0,1\}$ be an expression.
    Then, the conditional division $$v.\text{div}a\text{\_if}(s)\coloneqq 
    \begin{cases}
        \frac{v}{a} & s=1\\
        v   & s=0
    \end{cases}
    $$
    is an expression. Note that due to the restrictions on the arguments,
    \\$v.\text{div}a\text{\_if}(s)\equiv v - \relu((1-\frac{1}{a})v - (1-s))$, which is expressible by a neuron.
    \item [-] Let $0\leq v\leq1$ be a variable, let $0\leq x\leq 1$ be an expression, and let $s\in\{0,1\}$ be another expression. Then, the conditional assignment $$v.\text{set\_if}(s,x)\coloneqq 
    \begin{cases}
        x & s=1\\
        v & s=0
    \end{cases}
    $$
    \end{itemize}
    is an expression. Note that due to the restrictions on the arguments, 
    \\$v.set\_if(s,x)\equiv v.decrease\_if(s,v) + \relu(x-(1-s))$, which is expressible by 2 neurons in one layer and another neuron in the next layer.
\end{itemize}
\end{definition}

\subsubsection*{Constructing $F_L$}
\begin{lemma}    
    Let $a\in\Rat$, then there exist $d\in\Nat$ for which there is a $d$-dimension switched network
    $$F=\big((l_1,\ldots,l_m),\{(1,0),(0,1),(0,0)\}\times\Nat^2\times\Rat^8,$$$$(x_1,x_2)\rightarrow ax_1x_2^3\big)$$
    That is, there exists a switched network that for every $a\in\Rat,x_1,x_2\in\Nat$ computes $a x_1 x_2^3$.
\end{lemma}
\begin{proof}
Following the code in \Cref{lst:msg_len}, which conforms to \Cref{def:MLP_code}, it is not difficult to verify that it defines a $12$-dimension switched network $$F=\Big(\big((w_1,b_1),\ldots,(w_m,b_m)\big),\{(1,0),(0,1),(0,0)\}\times$$$$\Nat^2\times\Rat^8,(x_1,x_2)\rightarrow ax_1x_2^3\Big)$$ for some rational matrices and biases $(w_i,b_i)$.
\end{proof}

\subsubsection*{Constructing $F_{B2Q_1}$}
The sub-network $F_{B2Q_1}$ is identical to $F_{B2Q}$ which is described later. The difference is that $F_{B2Q_1}$ is used only once, in the initialization stage of the whole $R-GNN$ run. There, it converts the input feature, provided by the user in $\RB$ encoding, to $\RQ$ encoding, which is then used by $F_{I2QU}$.

\subsubsection*{Constructing $F_{I2QU}$}
\begin{lemma}
    Define $D\coloneqq \{(1,0),(0,1),(0,0)\}\times\Nat\times\RQ\times\Ratzo\times\{0\}\}$, and define $\forall x\in\Nat\;\forall y\in\RQ\;\;g(x,y)\coloneqq(\frac{1}{4^{x+1}}+\frac{y}{4^{x+2}}+\Sigma_{i\in[x]}\frac{3}{4^i})$. Then, there exists a switched network    
    $$F=((l_1,\ldots,l_m),D, g)$$
    That is, there exists a switched network that given $x\in\Nat$ and $y\in\RQ$ outputs the translation of $x$ to Unary Rational Quaternary concatenated with the digit '1' in $\RQ$ concatenated to $y$.
\end{lemma}
\begin{proof}
    Following the code in \Cref{lst:i2qu}, which conforms to \Cref{def:MLP_code}, it is not difficult to verify that it defines the required switched network.
\end{proof}

\subsubsection*{Constructing $F_{B2Q}$}
\begin{lemma}
    Define
    $$\forall L\in\Nat\;\; D_L\coloneqq \{(1,0),(0,1),(0,0)\}\times\{L\}\times$$$$\{x:x=\rbe(y),y\in\CB_L\}\times\Ratzo\times\Rat^{12}\}$$
    Then, there exists a recurrent MLP $F=(l_1,\ldots,l_m)$ such that 
    $\forall L\in\Nat$ it holds that $S_L=(F,D_L,\rqe\circ\rbe^{-1})$ is a switched network.
    That is, there exists a recurrent MLP such that given a message length $L\in\Nat$ and a rational binary encoding of a message $x\in\CB_L$, outputs the message's rational quaternary encoding.
\end{lemma}
\begin{proof}
    Following the code in \Cref{lst:b2q}, which conforms to \Cref{def:MLP_code}, it is not difficult to verify that it defines the required switched network.
\end{proof}

\subsubsection*{Constructing $F_{Q2B}$}
Unlike $F_{B2Q}$, $F_{Q2B}$ does not have the message length as input, since the $\RQ$ encoding allows us to identify, without receiving the length explicitly, when we have read all digits and the message is empty: The value of an empty message is $0$ while the value of any non-empty message is at least $\frac{1}{4}$.
\begin{lemma}
    Define $D\coloneqq \{(1,0),(0,1),(0,0)\}\times\RQ\times\Ratzo\times\{0\}\times\Ratzo\}$, then there exists a switched network    
    $$F=((l_1,\ldots,l_m),D,\rbe\circ\rqe^{-1})$$
    That is, there exists a switched network that for every $x\in\CB$ translates its rational quaternary encoding to its rational binary encoding.
\end{lemma}
\begin{proof}
    Following the code in \Cref{lst:q2b}, which conforms to \Cref{def:MLP_code}, it is not difficult to verify that it defines the required switched network.
\end{proof}

\subsubsection*{Constructing $F_R$}
\begin{lemma}
    Define $D\coloneqq \{(1,0),(0,1),(0,0)\}\times\RQ^2\times\Ratzo^2\times\{0\}\times\Ratzo\}$, then there exists a switched network 
    $$F=((l_1,\ldots,l_m),D,(1,\rbe\circ\rqe^{-1}))$$
    That is, there exists a switched network that for every $x\in\CB$ outputs the indication $1$ and the translation of $x$ from its rational quaternary encoding to its rational binary encoding.
\end{lemma}
\begin{proof}
    Following the code in \Cref{lst:QRes2BRes}, which conforms to \Cref{def:MLP_code}, it is not difficult to verify that it defines the required switched network.
\end{proof}

The last sub-network we describe, $F_S$, is not a switched network. It is 'on' in every recurrence, in parallel to whatever switched network is 'on' in that recurrence. Also, part of its input dimensions are external i.e. they do not assume their output-layer value from the previous recurrence, rather, they are set externally at the beginning of each recurrence. We formally define this kind of network as follows.
\begin{definition}\label{def:external_in_net}
    A $d$-dimension $m$-depth \emph{recurrent MLP with external inputs} \\$F=((l_1,\ldots,l_m),z_1,\ldots,z_L)$
    is a variation of a $d$-dimension recurrent MLP, such that the last $L$ dimensions are external input sequences $\{z_i^{(t)}\}_{i\in[L],t\in\Nat}$. That is,
      $$f_F^{(0)}(x)\coloneqq x\in\Rat^d,\;\;\forall t>0\;\;f_F^{(t)}(x)\coloneqq$$$$ f_F\Big(f_F^{(t-1)}(x)[1,d-L],\big(z_1(t),\ldots,z_L(t)\big)\Big)$$
\end{definition}

\subsubsection*{Constructing $F_S$}
The sub-network $F_S$ implements a mechanism that synchronizes the nodes' Turing-machine emulations. 
The idea is that at specific recurrences, all nodes record and broadcast their status. Then, for the next $|G|$ recurrences they consider their own recording and the messages they receive, and broadcast whether there is an indication of a non-finished node. This means that if and only if there is such node then all nodes will be aware of it at the end of those $|G|$ recurrences - because it takes at most $|G|$ recurrences for the information to propagate. In other words, all nodes have the same global status-snapshot at the end of those $|G|$ recurrences.

If according to the snapshot all nodes' Turing machines are finished, then all nodes start their new computation in the next recurrence - exactly at the same time. In addition, we construct the R-GNN such that when a node's Turing machine is finished the node continues to send the same result-message in subsequent recurrences until starting a new computation. The above scheme assures that all nodes' Turing machines start computation $i+1$ - emulating iteration $i+1$ of the emulated S-MP-GA algorithm - with input which contains the sum of their neighbors' $i^{th}$-computation results.

Due to technical limitations of computing with MLP, $F_S$ runs in cycles of $2|G|$ recurrences instead of $|G|$: The first $|G|$ recurrences are used to reset its state, and the second $|G|$ recurrences implement the scheme above. It has three external inputs: The graph size; the sum of neighbors' "some node not finished yet at the beginning of the cycle" indicator; a 'finished' indicator for the node's current computation, and two outputs: "should start new computation" - used by $F_{B2Q}$; and "some node not finished yet at the beginning of the cycle" indicator - to be sent to neighbors. The first output is $0$ throughout the cycle except for the last recurrence where it is $1$ in case all nodes are finished, and the second output is $0$ throughout the first half of the cycle -  where it is basically disabled - and becomes $1$ in the second half in case an indication of a non-finished node was received.

The following Lemma formalizes the described behavior.
\begin{lemma}
    Let $n\in\Nat$, then there exist a $d$-dimension recurrent with external inputs MLP 
    \\$F=((l_1,\ldots,l_m),z_1,z_2,z_3)$ such that:
    
    $$f_F^{(t)}(1)\coloneqq
    \begin{cases}
        0 & t\mod 2n > 0\\
        1-f_F^{(t-1)}(2) & t\mod 2n = 0
    \end{cases}$$

    $$f_F^{(t)}(2)\coloneqq
    \begin{cases}
        0 & t \textup{ mod } 2n \leq n\\
        \min(1,\Sigma^{t}_{i=t-(t\textup{ mod } n)}(z_2^{(i)}+z_3^{(i)})) & t \textup{ mod } 2n > n
    \end{cases}$$        
\end{lemma}
\begin{proof}
    Following the code in \Cref{lst:sync}, which conforms to \Cref{def:MLP_code}, it is not difficult to verify
    that it defines the required recurrent network with external inputs.
\end{proof}

\subsubsection*{Constructing The R-GNN $N=(A,F)$}\label{subsubsec:sgnn_construct}
We are now ready to put together all the sub-networks above into one recurrent MLP $F$. Their I/O dimension, initial values - corresponding to their part in the $A$ vector, and individual functionality, are already defined in their dedicated subsections above. Here we describe how they are connected to one another to form the whole recurrent network $F$. We refer to them by their names in the subsections describing their construction. Please refer to \Cref{fig:construction} for an illustration of the construction.

\begin{myremark}
Note that the positions of the inputs to the various sub-networks, as implied by the sub-networks' definitions and the construction below, do not match their positions according to the definition of $N^{(0)}(G,v)$ in \Cref{def:spiral_gnn}. The reason for the permutation is so both the definition of $N^{(0)}(G,v)$ and the description of the construction are easier to follow. Reversing the permutation e.g. by adding a prelude and postlude layers is trivial and for purpose of focus we do not include it.
\end{myremark}

For a $d$-dimension $m$-depth sub-network $F_H$ we define $d(F_H)\coloneqq d;m(F_H)\coloneqq m$.
Note that when referring to the dimensions of a sub-network we mean its I/O dimensions. Whatever are the dimensions and arguments of its hidden layers, they do not make a difference to our construction of $F$. Assume that $F_H$ is embedded in $F$ such that dimensions $a..(a+d(F_H)-1)$ in the input layer of $F$ are the input layer of $F_H$. We define 
$E(F_H)\coloneqq (a+d(F_H)-1)$ the last dimension of the part of $F_H$ in $F$.

Define $m_{\text{max}}\coloneqq max(m(F_H) \mid \text{H is a sub-network})$ the maximum depth over all sub-networks. We assume all sub-networks of lower depth are extended to depth $m_{\text{max}}$ by simply passing on their output layer, hence we can refer to layer $m_{\text{max}}$ of any sub-network. We denote by $y'_{F_H}(i)$ the $i^\th$ dimension in the output layer of the $F_H$ part of $F$, that is, the $a+i-1$ neuron in layer $m_{\text{max}}$ of $F$. To implement the interoperability between the sub-networks, we add additional layers after $y'$, up to the final output layer which we denote by $y$. The interconnections are defined by describing certain dimensions in $y$ as functions of (also) dimensions in other sub-networks' outputs and the aggregation values - in $y'$.
Dimensions in $y$ for which we do not define functions, simply assume their values in $y'$ as is.
We denote by $y_{F_H}(i)$ the $a+i-1$ neuron in the last layer of $F$ - whose output in one recurrence is the value of the $a+i-1$ input neuron in the following recurrence. Note that $y,y'$ are not to be confused with the $y,y'$ layers mentioned in the individual description of $F_M$. 

Let $H$ be the unique name of one of the sub-networks e.g. $H=B2Q$. We denote by $S_H$ the dimensions of $F$ that are the sum of the neighbors' $F_H$. For example, $S_M(1)$ is the sum, over all neighbors, of the first dimension of $F_M$'s output in the previous recurrence, and is in dimension $E(F_S)+E(F_{B2Q})+1$ in $F$. As described later, the sum values are used as inputs, only in the extra layers of $F$ - after the $y'$ layer. That is, they simply pass on from the input layer until the last layer, and some of their neurons are inputs to other parts of $F$ in the extra layers after $y'$.

And so, the whole flow is established in which at every recurrence of $F$ only specific sub-networks are active, and eventually their outputs become inputs for other sub-networks which in turn become the active ones etc.
\\\textbf{Subnetwork $F_L$}, dimensions $[1..d(F_L)]$
\\\textbf{Subnetwork $F_{B2Q_1}$}, dimensions $[d(F_L)+1..d(F_L)+d(F_{B2Q_1})]$
\\\myindent $y_{F_{B2Q_1}}(1)\coloneqq \relu\big(y'_{F_{B2Q_1}}(1)+y'_{F_L}(2)\big)$ i.e. $F_{B2Q_1}$ should start after $F_L$ is finished.
\\\myindent $y_{F_{B2Q_1}}(3)\coloneqq \relu\Big(y'_{F_L}(3) - \big(1-y'_{F_L}(2)\big)\Big) + \relu\Big(y'_{F_{Q2B_1}}(3) - \big(y'_{F_L}(2)\big)\Big)$ i.e. $F_{Q2B_1}$ feature length input is set to be the dimension with that data used also by $F_L$, for the beginning of a new computation, otherwise it maintains its value.
\\\textbf{Subnetwork $F_{I2UQ}$}, dimensions $[E(F_{B2Q_1})+1..E(F_{B2Q_1})+d(F_{I2UQ})]$
\\\myindent $y_{F_{I2UQ}}(1)\coloneqq \relu\big(y'_{F_{I2UQ}}(1)+y'_{F_{B2Q_1}}(2)\big)$ i.e. $F_{I2UQ}$ should start after $F_{B2Q_1}$ is finished.
\\\myindent $y_{F_{I2UQ}}(3)\coloneqq \relu\Big(y'_{F_L}(4) - \big(1-y'_{F_{B2Q_1}}(2)\big)\Big) + \relu\Big(y'_{F_{I2UQ}}(3) - \big(y'_{F_{B2Q_1}}(2)\big)\Big)$ i.e. $F_{I2UQ}$ graph-size input is set to be the dimension with that data used also by $F_L$, for the beginning of a new computation, otherwise it maintains its value.
\\\myindent $y_{F_{I2UQ}}(4)\coloneqq \relu\Big(y'_{F_{B2Q_1}}(5) - \big(1-y'_{F_{B2Q_1}}(2)\big)\Big) + \relu\Big(y'_{F_{I2UQ}}(4) - \big(y'_{F_{B2Q_1}}(2)\big)\Big)$ i.e. $F_{I2UQ}$ num-in-$\RQ$ input is set to be the output of $F_{B2Q_1}$, for the beginning of a new computation, otherwise it maintains its value.
\\\textbf{Subnetwork $F_{B2Q}$}, dimensions $[E(F_{I2UQ})+1..E(F_{I2UQ})+d(F_{B2Q})]$
\\\myindent $y_{F_{B2Q}}(1)\coloneqq \relu\big(y'_{F_{B2Q}}(1)+y'_{F_S}(1)-y'_{F_{I2UQ}}(1)\big)$ i.e. $F_{B2Q}$ should start only after $F_{I2UQ}$ has switched off, and from then on it should be on either if it is in the middle of a computation or if it received a signal from $F_S$ that it can start again. Note that a signal from $F_S$ means that  $F_M;F_{Q2B}$ are off.
\\\myindent $y_{F_{B2Q}}(3)\coloneqq y'_{F_L}(5)$ i.e. $F_{B2Q}$ should take $L_{n,k}$ from the first output of $F_L$.
\\\myindent $y_{F_{B2Q}}(4)\coloneqq \relu\big(y'_{S_M}(8)-y'_{F_S}(1)-y'_{F_{I2UQ}}(1)\big)$ i.e. $F_{B2Q}$ should take new input from the sum of neighbors (at position: second output of $F_M$) when it is signaled to start again.
\\\textbf{Subnetwork $F_M$}, dimensions $[E(F_{B2Q})+1..E(F_{B2Q})+d(F_M)]$
\\\myindent $y_{F_M}(1)\coloneqq \relu\big(y'_{F_M}(1) + y'_{F_{B2Q}}(2)\big)$ i.e. $F_M$ should turn on after $F_{B2Q}$ turns off.
\\\myindent $y_{F_M}(3)\coloneqq \relu\Big(y'_{F_M}(7) - \big(1-y'_{F_{B2Q}}(2)\big) + \relu\Big(y'_{F_M}(3) - \big(y'_{F_{B2Q}}(2)\big) \Big)$ i.e. $F_M$ first input is set to be the first output for the beginning of a new computation, otherwise it maintains its value.
\\\myindent $y_{F_M}(4)\coloneqq \relu\Big(y'_{F_{B2Q}}(5) - \big(1-y'_{F_{B2Q}}(2)\big) \Big) + \relu\Big(y'_{F_M}(4) - \big(y'_{F_{B2Q}}(2)\big) \Big)$ i.e. $F_M$ second input is set to be the output of $F_{B2Q}$ for the beginning of a new computation, otherwise it maintains its value.
\\\myindent $y_{F_M}(7)\coloneqq \relu\Big(y'_{F_{I2UQ}}(5) - \big(1-y'_{F_{I2UQ}}(2)\big) + \relu\Big(y'_{F_M}(7) - \big(y'_{F_{I2UQ}}(2)\big) \Big)$ i.e. $F_M$ first output gets the result of $F_{I2UQ}$ and maintains it as long as there is no computation, so when the first computation starts $y_{F_M}(3)$ will read the the result of $F_{I2UQ}$.
\\\textbf{Subnetwork $F_R$}, dimensions $[E(F_M)+1..E(F_M)+d(F_R)]$
\\\myindent $y_{F_R}(1)\coloneqq \relu\big(y'_{F_R}(1) + y'_{F_M}(2)\big)$ i.e. $F_R$ should turn on after $F_M$ turns off.
\\\myindent $y_{F_R}(3)\coloneqq \relu\Big(y'_{F_M}(10) - \big(1-y'_{F_M}(2)\big) \Big) + \relu\Big(y'_{F_R}(3) - \big(y'_{F_M}(2)\big)\Big)$ i.e. $F_R$ 'allFinished' input is set to be the last output of $F_M$ upon start.
\\\myindent $y_{F_R}(4)\coloneqq \relu\Big(y'_{F_M}(9) - \big(1-y'_{F_M}(2)\big) \Big) + \relu\Big(y'_{F_R}(4) - \big(y'_{F_M}(2)\big)\Big)$ i.e. $F_R$ 'final feature' input is set to be the one before last output of $F_M$ upon start.
\\\textbf{Subnetwork $F_{Q2B}$}, dimensions $[E(F_R)+1..E(F_R)+d(F_{Q2B})]$
\\\myindent $y_{F_{Q2B}}(1)\coloneqq \relu\big(y'_{F_{Q2B}}(1) + y'_{F_R}(2)-y'_{F_R}(5)\big)$ i.e. $F_{Q2B}$ should turn on after $F_R$ turns off but not if overall computation of the R-GNN is finished.
\\\myindent $y_{F_{Q2B}}(3)\coloneqq \relu\Big(y'_{F_M}(8) - \big(1-y'_{F_R}(2)\big)\Big) + \relu\Big(y'_{F_{Q2B}}(3) - \big(y'_{F_R}(2)\big)\Big)$ i.e. $F_{Q2B}$ input is set to be the second output of $F_M$ for the beginning of a new computation, otherwise it maintains its value.
\\\textbf{Subnetwork $F_S$}, dimensions $[E(F_{Q2B})+1..E(F_{Q2B})+d(F_S)]$
\\\myindent $y_{F_S}(d(F_S)-2)\coloneqq y_{F_L}(2)$ i.e. $F_S$ first input is the graph size, which is found constantly in the second dimension of $F_L$.
\\\myindent $y_{F_S}(d(F_S)-1)\coloneqq\Big(1-$\\$\big(y'_{F_{B2Q}}(1)+y'_{F_M}(1)+y'_{F_{Q2B}}(1)+y'_{F_{B2Q}}(2)+y'_{F_M}(2)+y'_{F_{Q2B}}(2)+y'_{F_R}(2)\big)\Big)$ i.e. $F_S$ second input should be $1$ if the whole computation process is currently off i.e. most recent computation is finished. The reason for sampling both the "switch on" and "switch just turned off" dimensions is to avoid getting into the analysis of corner-case-timing cases.
\\\myindent $y_{F_S}(d(F_S))\coloneqq y'_{S_S}(2)$ i.e. $F_S$ third input should be the sum of the neighbors' $F_S$ indication if it has received a "not-finished" signal.
\end{proof}


\newpage
\lstset{style=sharpc}
\onecolumn
\begin{lstlisting}[label={lst:mpgc_to_mpgc_one},columns=fullflexible,caption={Emulate MP-LGA}, frame=lines,escapeinside={(*}{*)}, numbers=left,breaklines=true, postbreak=\mbox{\textcolor{red}{$\hookrightarrow$}\space},basicstyle=\footnotesize,  commentstyle=\color{commentsColor}]
    
    Initialize:  // impelmentation of (*$C^{(0)}(G,v)$*)
        dim1.graph_size = graph_size
        dim1.feature = feature
        dim1.disabled = false
        dim1.tmp_disabled = false                
        dim1.neighbors_values_and_counts = {}        
        dim1.inner_loop_counter = 0
        dim1.outer_loop_counter = 0                
        dim1.receive_max = 0
        dim1.MP_GC_dim1 = MP_GC_init_dim1
        dim1.MP_GC_iteration_count = 0
        dim2.count = 1
        dim2.value = feature
        dim3 = 0
        dim4 = 0

    run(prev_dim1, neighbors_dim2_sum, prev_dim3){                
        output_dim1 = prev_dim1.copy() // start with a copy, then set what needs to be updated
        output_dim3 = prev_dim3 // used to indicate finishing the overall computation, start with same value as previous
        output_dim4 = prev_dim4 // used to hold the final value when the computation is finished
        if(prev_dim1.MP_GC_iteration_count == prev_dim1.graph_size+1){// finished whole computation, output_dim2.value should have the final output            
            output_dim3 = 1
            output_dim4 = prev_dim1.feature
        }
        else if(prev_dim1.outer_loop_counter == prev_dim1.graph_size){// finished collecting multiset of neighbors' features, run MP_GC_func
            (MP_GC_output_dim1, MP_GC_output_dim2) = MP_GC_func(prev_dim1.MP_GC_dim1, prev_dim1.neighbors_values_and_counts)
            output_dim1.MP_GC_dim1 = MP_GC_output_dim1
            output_dim1.feature = MP_GC_output_dim2
            output_dim1.MP_GC_iteration_count = prev_dim1.MP_GC_iteration_count + 1
            output_dim1.neighbors_values_and_counts = {}
            output_dim1.inner_loop_counter = 0                    
            output_dim1.outer_loop_counter = 0                    
        }                
        else if(prev_dim1.outer_loop_counter < prev_dim1.graph_size){// still collecting            
            if(prev_dim1.inner_loop_counter==prev_dim1.graph_size^2){ // finished isolating max among uncollected    
                send_if_max(prev_dim1, output_dim1, output_dim2)
                output_dim1.receive_max = 1 // next stage is to read neighbors that sent max
                output_dim1.inner_loop_counter = 0
            }
            else if(prev_dim1.receive_max == 1){
                receive_max(prev_dim1, neighbors_dim2_sum, output_dim1, output_dim2)
                output_dim1.receive_max = 0                 
                output_dim1.outer_loop_counter +=1                        
            }
            else{
                find_max(prev_dim1, neighbors_dim2_sum, output_dim1, output_dim2)
                output_dim1.inner_loop_counter +=1                        
            }                    
        }            
    }

    find_max(prev_dim1, neighbors_dim2_sum, output_dim1, output_dim2){
        neighbors_avg_value = neighbors_dim2_sum.value / neighbors_dim2_sum.count
        output_dim=2 = 1
        // if the vertex is disabled (or tmpDisabled) or its initial feature is lower than the observed value, then propogate the observed value
        output_dim2.value = neighbors_avg_value
        if (prev_dim1.feature < neighbors_avg_value)
        {
            // vertex value is lower, then tmpDisable vertex so eventually the only non-tmpDisabled vertices will be those with maximum value among the enabled vertices in the graph.        
            output_dim1.tmp_disabled = true          
        }
        else if (!prev_dim1.disabled && !prev_dim1.tmp_disabled)
        {                    
            // if the vertex is enabled and its initial feature higher than the observed value then propogate its value
            output_dim2.value = prev_dim1.feature                        
        }                
    }
    
    send_if_max(prev_dim1, output_dim1, output_dim2){
        if (!prev_dim1.disabled && !prev_dim1.tmp_disabled)
        {  // vertex is one of those with max value among the enabled, send its value to its neighbors, and disable it
            output_dim1.disabled = true    
            output_dim2.count = 1    
            output_dim2.value = prev_dim1.feature                                         
        }
        else
        {  // vertex is either disabled because it already sent its (relatively high) value, or it is 
        // tmpDisabled because of its relatively low value. Then, signal that its shouldn't be counted                    
            output_dim2.count = 0
            output_dim2.value = 0                                
        }                            
    }
    
    receive_max(prev_dim1, neighbors_dim2_sum, output_dim1, output_dim2){                                
        neighbors_avg_value = neighbors_dim2_sum.value / neighbors_dim2_sum.count
        // we assume that at this point the vertices that sent a non-zero value, and '1' dim2.count, all share  the same value - the maximum value among non-disabled vertices in the graph. Hence, their average is that maximum value.
        if (neighbors_avg_value > 0)
        // otherwise the vertex has no neighbors with the max value, since we assume non-zero initial values
        {             
            output_dim1.neighbors_values_and_counts.add(neighbors_dim2_sum.count, neighbors_avg_value)
        }                    
        if (!prev_dim1.disabled)
        // vertex still hasn't got to be a max non-disabled value, hence it continues to try - until all higher values will be recorded.
        {
            output_dim1.tmp_disabled = false
            output_dim2.count = 1
            output_dim2.value = prev_dim1.initial_feature                     
        }
        else
        {
            output_dim2.count = 0
            output_dim2.value = 0                    
        }
    }    
\end{lstlisting}

\begin{lstlisting}[label={lst:msg_len},columns=fullflexible,caption={Implement Message Length Computation}, frame=lines,escapeinside={(*}{*)}, mathescape, numbers=left,breaklines=true, postbreak=\mbox{\textcolor{red}{$\hookrightarrow$}\space},basicstyle=\footnotesize,  commentstyle=\color{commentsColor}]

    // Computes $3\cdot$s_initFeatLen$\cdot$s_graphSize$^4$
    
    Initialize:    
        [1] s_switchOn = 1
        [2] s_switchTurnedOff = 0
        [3] s_initFeatLen = 0
        [4] s_graphSize = graphSize
        [5] s_result = 0            
        [6] s_counter1 = 0
        [7] s_counter2 = 0
        [8] s_counter3 = 0
        [9] s_counter4 = 0
        [10] s_stateReset1 = 0
        [11] s_stateReset2 = 0
        [12] s_stateReset3 = 0
        [13] s_stateAddTo1 = 0        
        
    RecurrentOp:                
        var prevSwitchVal = s_switchOn
        var edgeCaseOne = Lsig(1 - (s_input1 - 1)) // s_input1 = 1
        // we add 3 times becaues we want to multiply by 3
        s_result.icrease_if$(\lsig($s_switchOn - s_stateReset1$ - $s_stateReset2 $- $s_stateReset3 $),1)$
        s_result.icrease_if$(\lsig($s_switchOn - s_stateReset1$ - $s_stateReset2 $- $s_stateReset3 $),1)$
        s_result.icrease_if$(\lsig($s_switchOn - s_stateReset1$ - $s_stateReset2 $- $s_stateReset3 $),1)$
        
        var goodOne $= \lsig(\lsig($s_graphSize $-$ s_counter1$)-(1-$s_stateAddTo1) - edgeCaseOne) // we want to add and we can
        s_stateReset1 $= \lsig($s_stateReset1$ - \lsig(1 - $s_counter1$))$ // maintain
        s_stateReset1 $= \lsig($s_stateReset1$ + \lsig($s_stateAddTo1$ - \lsig($s_graphSize$ - 1 - $s_counter1$)))$ // turn on: we want to add counter reached limit, reset counter
        s_stateReset1 $= \lsig$(s_stateReset1 -edgeCaseOne)
        s_stateAddTo1 = \lsig(s_stateAddTo1 - s_stateReset1-edgeCaseOne);
        s_counter1.decrease_if$($s_stateReset1, $1)$ // resetting
        s_counter1.increase_if$($goodOne$, 1)$     
        
        var addTo2 $= \lsig(\lsig($s_stateReset1 $- \lsig($s_counter1$)))$ // reset1 finished
        var goodTwo $= \lsig(\lsig($s_graphSize $-$ s_counter2$) - (1 - $addTo2$))$ // we want to add and we can
        s_stateReset2 $= \lsig($s_stateReset2 $- \lsig(1 - ($s_counter2$)))$ // maintain
        s_stateReset2 = \lsig(s_stateReset2 + \lsig(addTo2 - \lsig(s_graphSize - 1 - s_counter2))); // turn on: we want to add but cannot
        s_counter2.decrease_if(s_stateReset2, 1)
        s_counter2.increase_if(goodTwo, 1)
        
        var addTo3 $= \lsig(\lsig($s_stateReset2 $- \lsig($s_counter2$)))$ // reset2 finished
        var goodThree $= \lsig(\lsig($s_graphSize $- 1 - $s_counter3$) - (1 - $addTo3$))$ // we want to add and we can
        s_stateReset3 $= \lsig($s_stateReset3 $- \lsig(1 - ($s_counter3$)))$ // maintain
        s_stateReset3 $= \lsig($s_stateReset3 $+ \lsig($addTo3 $- \lsig($s_graphSize $- 1 - $s_counter3$)))$ // turn on: we want
        s_counter3.decrease_if(s_stateReset3, 1)
        s_counter3.increase_if(goodThree, 1)        

        var addTo4 $= \lsig(\lsig($s_stateReset3 $- \lsig($s_counter3$)))$ // reset3 finished
        var goodFour $= \lsig(\lsig($s_graphSize $- 1 - $s_counter4$) - (1 - $addTo4$))$ // we want to add and we can
        s_stateReset4 $= \lsig($s_stateReset4 $- \lsig(1 - ($s_counter4$)))$ // maintain
        s_stateReset4 $= \lsig($s_stateReset4 $+ \lsig($addTo4 $- \lsig($s_graphSize $- 1 - $s_counter4$)))$ // turn on: we want
        s_counter4.decrease_if(s_stateReset3, 1)
        s_counter4.increase_if(goodThree, 1)        
        
        var addTo5 = \lsig(\lsig(s_stateReset4 - \lsig(s_counter4))); // reset3 finished
        var goodFive $= \lsig(\lsig($s_initFeatLen $- 1 - $s_counter5$) - (1 - $addTo5$))$ // we want to add and we can
        s_counter5.increase_if(goodFive, 1)
        
        s_switchOn $= 1-\lsig(\lsig($s_counter5 $-$ s_initFeatLen $ + 2) + $addTo5 $+ \lsig(1 - $goodFive$) - 2)$
        s_stateAddTo1 $= \lsig(1 - $s_stateReset1 $-$ s_stateReset2 $-$ s_stateReset3$-$ s_stateReset4 - (1-s_switchOn))
        
        s_switchTurnedOff = $\relu($prevSwitchVal-s_switchOn$)$        
            
\end{lstlisting}

\begin{lstlisting}[label={lst:i2qu},columns=fullflexible,caption={Implement I2QU Translation + Concatenation To Other}, frame=lines,escapeinside={(*}{*)}, mathescape, numbers=left,breaklines=true, postbreak=\mbox{\textcolor{red}{$\hookrightarrow$}\space},basicstyle=\footnotesize,  commentstyle=\color{commentsColor}]

    Initialize:
        [1] s_switchOn = 0
        [2] s_switchTurnedOff = 0
        [3] s_numberLeftToProcess = 0
        [4] s_otherInRQ = 0
        [5] s_result = 0
        [6] s_stateAddToNumber = 0
        
    RecurrentOp:        
        var prevSwitchVal = s_stateAddToNumber
        s_stateAddToNumber = $\lsig$(s_numberLeftToProcess + s_switchOn-1))
        s_switchOn = s_stateAddToNumber
        s_switchTurnedOff = $\relu$(prevSwitchVal-s_switchOn)
        var switchTurnedOn = $\relu$(s_switchOn-prevSwitchVal)

        // init procedure, when switch turns on                
        s_result.set_if(switchTurnedOn, s_otherInRQ)        
        s_result.div4_if(switchTurnedOn) // together with next line: insert a separating "0" i.e. 1/4 in RQ 
        s_result.increase_if(switchTurnedOn, 1/4.0)

        // operation in add to number state    
        s_result.div4_if(s_stateAddToNumber)
        s_result.increase_if(s_stateAddToNumber, 3/4.0)
        s_numberLeftToProcess.decrease_if(s_stateAddToNumber, 1)        
        
\end{lstlisting}

\begin{lstlisting}[label={lst:b2q},name=b2q,columns=fullflexible,caption={Implement B2Q Translation}, frame=lines,escapeinside={(*}{*)}, mathescape, numbers=left,breaklines=true, postbreak=\mbox{\textcolor{red}{$\hookrightarrow$}\space},basicstyle=\footnotesize,  commentstyle=\color{commentsColor}]

    /* The idea of the algorithm in general lines is as follows:
     Initially, $x = \Sigma_{i\in[m]}\frac{a_i}{2^i}$ where $x=$s_number_in_process, $m=$s_messageBitLength.
     All relevant variables are reset to their starting values in the s_stateInit stage. Then,
     For i=1..m
        Assume we are left with $x = \Sigma_{j\in[i..m]}\frac{a_j}{2^j}$. The s_stateReduce stage implements:
        x = $\relu(\Sigma_{j\in[i+1..m}\frac{1}{2^j})$ // at that point $a_i=1\Rightarrow x\geq \frac{1}{2^m}$ and $a_i=0\Rightarrow x\leq0$        
        Then the s_stateShiftLeft stage implements:
        $x = \lsig(2^mx)$ // at that point $a_i=1\Rightarrow x=1$ and $a_i=0\Rightarrow x=0$        
        Then: the s_addToNumber stage updates the result accordingly - adding $\frac{1}{4^i}$ or $\frac{3}{4^i}$, and so is the s_numberLeftToProcess.            
    */  

    Initialize:
        [1] s_switch = 0
        [2] s_switchTurnedOff = 0      
        [3] s_messageBitLength = 0            
        [4] s_numberLeftToProcess = 0      
        [5] s_numberInC4 = 0      
        [6] s_number_in_process = 0                        
        [7] s_stateInit = 0
        [8] s_stateReduce = 0     
        [9] s_stateShiftLeft = 0
        [10] s_stateAddToNumber = 0
        [11] s_maxDigitsToTheRight = 0
        [12] s_digitsToTheLeft = 0
        [13] s_digitsToTheRight = 0
        [14] s_nextReduce = 0        
        [16] s_C41 = 0
        [17] s_C43 = 0
        
    RecurrentOp:        
        var switchWasOff = $\lsig$(s_stateReduce + s_stateShiftLeft + s_stateAddToNumber + s_stateInit)
        var switchTurnedOn = $\relu$(s_switch-switchWasOff)
        s_stateInit = switchTurnedOn
        var prevSwitchVal = s_switch

        // determine state of current pass
        s_stateReduce $= \lsig(\lsig($s_stateReduce$ - \lsig(1 - $s_digitsToTheRight$)) + $\lsig($s_stateInit$ - \lsig($s_messageBitLength$ - s_maxDigitsToTheRight$))-(1-s_switch))$
        s_stateShiftLeft$ = \lsig(1 - $s_stateReduce$ - $s_stateInit$-(1-s_switch))$
        s_stateShiftLeft $= \lsig($s_stateShiftLeft $- \lsig(1 - $s_digitsToTheLeft)-(1-s_switch))
        s_stateAddToNumber$ = \lsig(\lsig($s_maxDigitsToTheRight$) - ($s_stateReduce$ +$ s_stateShiftLeft$ +$ s_stateInit)-(1-s_switch))
        
        s_stateInit $= \lsig(\lsig($s_messageBitLength$ -$ s_maxDigitsToTheRight$) - (1 -$ s_stateInit)-(1-s_switch))
        // operation in init mode
        var change$ = \lsig($s_messageBitLength$ -$ s_digitsToTheRight)
        s_digitsToTheRight.increase_if(s_stateInit, change)
        s_maxDigitsToTheRight.increase_if(s_stateInit, 1)
        s_nextReduce.set_if(s_stateInit, 1 / 4.0)
        s_C41.set_if(s_stateInit, 1 / 4.0)
        s_C43.set_if(s_stateInit, 3 / 4.0)
        s_numberInC4.set_if(s_stateInit, 0)
        
        // operation in the reduce state
        s_digitsToTheRight.decrease_if(s_stateReduce, 1)
        s_digitsToTheLeft.increase_if(s_stateReduce, 1)            
        var tmp = s_number_in_process
        tmp.decrease_if(s_stateReduce, s_nextReduce)
        s_number_in_process $= \lsig$(tmp)
        s_nextReduce.div2_if(s_stateReduce)
        
        // operation in the shift left state
        var tmp2 = s_number_in_process
        tmp2.increase_if(s_stateShiftLeft, s_number_in_process)
        s_number_in_process $= \lsig$(tmp2)
        s_digitsToTheLeft.decrease_if(s_stateShiftLeft, 1)
        s_nextReduce.increase_if(s_stateShiftLeft, s_nextReduce)
        s_digitsToTheRight.increase_if(s_stateShiftLeft, $\lsig($s_maxDigitsToTheRight$ - 1 -$ s_digitsToTheRight)        

        // operation in add to number state
        // multiplying s_number_in_process by 2, this is sometimes required to make it $\geq 1$ (all when the extracted digit is 1)
        var tmp3 = s_number_in_process
        tmp3.increase_if(s_stateAddToNumber, s_number_in_process)
        s_number_in_process $= \lsig$(tmp3)
        s_numberInC4.increase_if(s_stateAddToNumber, $\lsig$(s_C41 - s_number_in_process$)+ \lsig($s_C43 $- (1 - $s_number_in_process$)))$
        s_C41.div4_if(s_stateAddToNumber)
        s_C43.div4_if(s_stateAddToNumber)
        s_numberLeftToProcess.decrease_if(s_stateAddToNumber, $\lsig(0.5 - (1 - $s_number_in_process$)))$
        s_numberLeftToProcess = $\lsig$(s_numberLeftToProcess)
        s_numberLeftToProcess.increase_if(s_stateAddToNumber, s_numberLeftToProcess)
        s_number_in_process.decrease_if(s_stateAddToNumber, s_number_in_process)
        s_number_in_process.increase_if(s_stateAddToNumber, s_numberLeftToProcess)
        
        s_maxDigitsToTheRight.decrease_if(s_stateAddToNumber, 1)
        s_stateReduce.increase_if(s_stateAddToNumber, s_stateReduce)
        
        s_switch = $\lsig$(s_stateReduce + s_stateShiftLeft + s_stateAddToNumber + s_stateInit)
        s_switchTurnedOff = $\lsig$(prevSwitchVal-s_switch)
                
\end{lstlisting}

\begin{lstlisting}[label={lst:q2b},columns=fullflexible,caption={Implement Q2B Translation}, frame=lines,escapeinside={(*}{*)}, mathescape, numbers=left,breaklines=true, postbreak=\mbox{\textcolor{red}{$\hookrightarrow$}\space},basicstyle=\footnotesize,  commentstyle=\color{commentsColor}]

    Initialize:
        [1] s_switchOn = 0
        [2] s_switchTurnedOff = 0
        [3] s_numberLeftToProcess = 0
        [4] s_numberInBinary = 0
        [5] s_stateAddToNumber = 0
        [6] s_nextPosBinaryValue = 0
        
    RecurrentOp:        
        var prevSwitchVal = s_stateAddToNumber
        s_stateAddToNumber = \lsig($\lsig(8 \cdot $s_numberLeftToProcess$ - 1)-\relu(1-s_switchOn))$
        s_switchOn = s_stateAddToNumber
        s_switchTurnedOff = \relu(prevSwitchVal-s_switchOn)
        var switchTurnedOn = \relu(s_switchOn-prevSwitchVal)

        \\ init procedure, when switch turns on
        s_nextPosBinaryValue.set_if(switchTurnedOn, 0.5)
        s_numberInBinary.set_if(switchTurnedOn, 0)                
        
        \\ number translation procedure
        var extractedDigit = $\lsig(4 \cdot $ s_numberLeftToProcess$-2)$ // first digit is in {1,3} and we translate to {0,1}
        s_numberInBinary.increase_if(s_stateAddToNumber, \lsig(s_nextPosBinaryValue - (1 - extracted)))
        s_numberLeftToProcess.decrease_if(s_stateAddToNumber, $(1 + 2 \cdot$ extracted$)/4.0$)
        // next 2 lines essentially multiply by 4
        s_numberLeftToProcess = ChangeIfState(s_stateAddToNumber, s_numberLeftToProcess) // multiply by 2
        s_numberLeftToProcess = ChangeIfState(s_stateAddToNumber, s_numberLeftToProcess) // multiply by 2        
        s_nextPosBinaryValue = div2_if_state(s_stateAddToNumber)
            
\end{lstlisting}

\begin{lstlisting}[label={lst:QRes2BRes},columns=fullflexible,caption={Implement Translation Of Final Result}, frame=lines,escapeinside={(*}{*)}, mathescape, numbers=left,breaklines=true, postbreak=\mbox{\textcolor{red}{$\hookrightarrow$}\space},basicstyle=\footnotesize,  commentstyle=\color{commentsColor}]

    Initialize:
        [1] s_switchOn = 0       
        [2] s_switchTurnedOff = 0
        [3] s_allFinished = 0
        [4] s_numberLeftToProcess = 0
        [5] s_allFinishedZeroOne = 0
        [6] s_numberInBinary = 0
        [7] s_stateAddToNumber = 0
        [8] s_nextPosBinaryValue = 0
        [9] s_lockResult = 0 // once we have the final result (in dimensions 5,6) we want it to stay no matter what happens in the network.
        
    RecurrentOp:        
        var prevSwitchVal = s_stateAddToNumber
        var allFinishedZeroOne = $\relu(4\cdot$ s_allFinished$-2)$ // from RQ to {0,1}
        var switchOnAndAllFinished = s_switchOn+allFinishedZeroOne$-1$        
        s_stateAddToNumber = $\relu(\lsig(\lsig(8 \cdot $s_numberLeftToProcess$ - 1)-(1-switchOnAndAllFinished))-s_lockResult)$
        s_switchTurnedOff = $\relu($prevSwitchVal$-$s_switchOn $+$
                                            $\relu((1-$prevSwitchVal$)+($s_switchOn-switchOnAndAllFinished$)-1))$
        s_allFinishedZeroOne.set_if$(\lsig(\relu($s_switchTurnedOff + allFinishedZeroOne-1)+s_lockResult), 1)
        s_lockResult = \relu(s_lockResult + s_allFinishedZeroOne)        
        s_switchOn = s_stateAddToNumber
        var shouldInit = $\relu($s_switchOn$-$prevSwitchVal$)$        

        \\ init procedure, when switch turns on
        s_nextPosBinaryValue.set_if(shouldInit, 0.5)
        s_numberInBinary.set_if(shouldInit, 0)                
        
        \\ number translation procedure
        var extractedDigit = $\lsig(4 \cdot $ s_numberLeftToProcess$-2)$ // first digit is in {1,3} and we translate to {0,1}
        s_numberInBinary.increase_if(s_stateAddToNumber, \lsig(s_nextPosBinaryValue - (1 - extracted)))
        s_numberLeftToProcess.decrease_if(s_stateAddToNumber, $(1 + 2 \cdot$ extracted$)/4.0$)
        // next 2 lines essentially multiply by 4
        s_numberLeftToProcess = ChangeIfState(s_stateAddToNumber, s_numberLeftToProcess) // multiply by 2
        s_numberLeftToProcess = ChangeIfState(s_stateAddToNumber, s_numberLeftToProcess) // multiply by 2        
        s_nextPosBinaryValue = div2_if_state(s_stateAddToNumber)
            
\end{lstlisting}

\begin{lstlisting}[label={lst:sync},columns=fullflexible,caption={Implement Synchronizer ($F_S$)}, frame=lines,escapeinside={(*}{*)}, mathescape, numbers=left,breaklines=true, postbreak=\mbox{\textcolor{red}{$\hookrightarrow$}\space},basicstyle=\footnotesize,  commentstyle=\color{commentsColor}]

    Initialize:                
        [1] s_stateReadyForNextAlgoIteration = 0 // signals the initiation of a new computation - starting from B2Q
        [2] s_foundNotFinished = 0    
        [3] s_syncCountdown = 0        
        [4] s_stateSyncInProgress = 0
        [5] s_stateCountdownOver = 1
        [6] s_stateResetCountdown = 0              
        [7] s_syncNumOfCycles = 0 // external input, should be constant
        [8] s_otherNodesNotFinished = 0 // external input
        [9] s_curNodeFinished = 0 // external input
        
    RecurrentOp:
        s_stateReadyForNextAlgoIteration = 0; // updated during the pass
        var cannotStart $= \relu(1-$s_syncNumOfCycles$)$ // as long as s_syncNumOfCycles is not received, cannot start
        var startedResetAndNotFinished = $= 
              \lsig($ s_stateResetCountdown $ + \lsig($ s_syncRoundsCount$ - $s_syncCountdown$) - 1)$
        s_stateResetCountdown $= \relu($startedResetAndNotFinished$-$cannotStart$)$ // do nothing until can start 
        
        var countdownGEzeroNotInReset $= \lsig(\lsig($ s_syncCountdown$)- $s_stateResetCountdown$)$
        s_stateSyncInProgress $= \relu($countdownGEzeroNotInReset-cannotStart) // do nothing until can start 
        s_stateCountdownOver $= \relu\Big(\big(1-\lsig($s_syncCountdown$)\big)-$cannotStart$)\Big)$ // ...
                
        s_foundNotFinished = $\relu\Big(\lsig($s_foundNotFinished $+$s_otherNodesNotFinished $+ (1-$s_curNodeFinished$) - $s_stateResetCountdown$)-$cannotStart$\Big)$ // either already had not-finished indication in this cycle, or received indication of a non-finished node, or current node is not finished
        s_syncCountdown.decrease_if(s_stateSyncInProgress, 1) // if during sync, countdown
   
        s_stateReadyForNextAlgoIteration $= \lsig($s_stateCountdownOver$ + (1-$s_foundNotFinished$)-1)$ // will be 0 while cannotStart
        s_stateResetCountdown $=$ s_stateReadyForNextAlgoIteration$)$ // reached 0 and all finished, should start reset       
        s_foundNotFinished $= \lsig($s_foundNotFinished$ - $s_stateResetCountdown$)$ // reset also resets foundNotFinished
        s_syncCountdown.increase_if(s_stateResetCountdown, 1) // in reset mode continue to increase countdown        
\end{lstlisting}

\twocolumn
\section{Further Results}

\begin{theorem}\label{thm:disco}
    There exists an mp-invariant graph embedding $F:\CG_\CB\rightarrow\CB$ such that for every R-GNN $N$ there exists a disconnected graph $G$ for which $N(G)\neq F(G)$.
\end{theorem}
\begin{proof}
For all $(m,n)\in\Nat$ with $m,n\ge 3$, we define a graph $G_{m,n}$ to be the disjoint union of a cycle $C_m^0$ of length $m$ in which all nodes have initial feature $0$, and a cycle $C_n^1$ of length $n$ in which all nodes have the initial feature $1$. 

We define a function $F:\CG_\CB\to\CB$ by 
\[
F(G_{m,n})\coloneqq\begin{cases}
    1&\text{if }m\text{ is even}\\
    0&\text{if }m\text{ is odd}
\end{cases}
\]
for all $m,n\ge 3$ and $F(G)\coloneqq 0$ if $G$ is not isomorphic to some $G_{m,n}$ for $m,n\ge 3$. Clearly, $F$ is computable and mp-invariant.

Suppose for contradiction that there is a graph-level R-GNN $N$ computing $F$. We consider the computation of $N$ on a graph $G_{m,n}$. After the computation stops, all vertices $v\in V(C_m^0)$ will have the same feature vector $\boldsymbol x_{m+n}\in\mathbb Q^k$, and all vertices $w\in V(C_n^1)$ will have the same feature vector $\boldsymbol y_{m+n}\in\mathbb Q^k$. Here $k$ is a constant only depending on $N$. The vectors $\boldsymbol x_{m+n}$ and $\boldsymbol y_{m+n}$ may depend on the order $m+n$ of the input graph $G_{m,n}$, but not on $m$ and $n$ individually. To compute the final output, $N$ passes the aggregated vector $\frac{m\boldsymbol x_{m+n}+n\boldsymbol y_{m+n}}{m+n}$ as input to an MLP which will compute the output $M(\frac{m\boldsymbol x+n\boldsymbol y}{m+n})=\rbe(F_N(G_{m,n}))$.

Every MLP with ReLu activations computes a piecewise linear function. This means that we can partition $\mathbb Q^k$ into finitely many convex polytopes $Q_1,\ldots,Q_q$, and on each $Q_j$ the restriction of the function $M$ computed by our MLP is linear (see, for example, \cite{GroheSSB25}). Within each $Q_i$, there is an affine and hence convex subset $R_i$ where $M$ is $1$ (possibly, $R_i$ is empty). Thus there are finitely many convex subsets $R_1,\ldots,R_q\subseteq \mathbb Q^k$ such that for all $\boldsymbol z\in\mathbb Q^k$ we have
\[
M(\boldsymbol z)=1/2\quad\iff\quad\boldsymbol z\in\underbrace{\bigcup_{i=1}^q R_i}_{\eqqcolon R}.
\]
Let $\ell\in\mathbb N$ such that $\lfloor\frac{\ell-6}{2}\rfloor>q$. For $3\le m\le \ell-3$, let $\boldsymbol z_m\coloneqq \frac{m\boldsymbol x_\ell+(\ell-m)\boldsymbol y_\ell}{l}$. On input $G_{m,\ell-m}$, the output of $N$ is $\rbe^{-1}(M(\boldsymbol z_m))$. As $F(G_{m,n})=1\iff m$ is even, this means that $\boldsymbol z_m\in R$ for all even $m$ and $\boldsymbol z_m\not\in R$ for all odd $m$. Since $R$ is the union of $q$ sets $R_i$ and there are more than $q$ even $m$ between $3$ and $\ell-3$, there are an $i\in[q]$ and even $m_1<m_2$ such that $\boldsymbol z_{m_1},\boldsymbol z_{m_2}\in R_i$. However, $\boldsymbol z_{m_1+1}$ is a convex combination of $\boldsymbol z_{m_1}$ and $\boldsymbol z_{m_2}$, and thus $\boldsymbol z_{m_1+1}\in R_i\subseteq R$. It follows that $N(G_{m_1+1,\ell-m_1-1})=\rbe^{-1}(M(\boldsymbol z_{m+1}))=1$, while $F(G_{m_1+1,\ell-m_1-1})=0$.
\end{proof}

\thmgraphembedding*
\begin{proof}
    Define a feature transformation $F_{\text{T}}:\mathcal{CG}_\CB\to\CZ_\CB,\ F_{\text{T}}(G)(v)\coloneqq F(G)$. The connectivity of the input, together with \Cref{lem:nl2gl}, mean that the (graph embedding) mp-invariance of $F$ implies (feature transformation) mp-invariance of $F_{\text{T}}$. By the latter and \Cref{theo:main} we have that there is an R-GNN $N'$ such that $\forall G\in \mathcal{CG}_\CB\ \forall v\in V(G)\; N'(G,v)=F_{\text{T}}(G)(v)$. Let $N=(N',x\mapsto x)$ be a graph-embedding R-GNN where the final MLP computes the identity function, then we have that
    $$\forall G\in \mathcal{CG}_\CB\;N(G)=$$$$\rbe^{-1}\Big(
    \frac{1}{|G|}\Sigma_{v\in V(G)}\rbe(N'(G,v))\Big)=$$
    $$\rbe^{-1}\Big(\rbe\big(F_{\text{T}}(G)(v)\big)\Big)=F_{\text{T}}(G)(v)=F(G)$$
\end{proof}

\corrandominit*
\begin{proof}        
    We can view the computation of an R-GNN with RNI as a two-stage process: Given a graph $G\in\CG_{\CB}, G=\{V(G),E(G),\CB,Z(G)\}$, we first extend the initial feature of every node by a random number. This gives us a graph which we denote by $\tilde G$, $\tilde G = \{V(G),E(G),\CB,\tilde Z(G)\}$, with the same structure as $G$ but extended features.
    Then, we run a deterministic R-GNN on $\tilde G$.
    As the features of an R-GNN are rational numbers - finite precision, we restrict the length of the binary representation of the random numbers to $3\log n$, that is, we choose a random bitstring of length $3\log n$ for each node. 
    With probability greater than $2/3$, each node will get a different random number assigned to it, in which case $\tilde G$ is considered \emph{individualized}.
    
    Let $G\in\mathcal{CG}_\CB$ be a connected graph, and let $v\in V(G)$. For $\tilde G,v$ we define
    $$\tilde F(\tilde G)(v)\coloneqq 
    \begin{cases}
    F(G)(v)    & \tilde G \text{ is individualized}\\
    \text{null-value}           &\text{otherwise}
    \end{cases}
    $$
    Note that for any two augmented graphs $\tilde G, \tilde H$ that are connected and individualized, and two vertices $v\in V(\tilde G), v'\in V(\tilde H)$, it holds that $\mpc_{\tilde G}(v)=\mpc_{\tilde H}(v')\Rightarrow$ there is an isomorphism $f:V(\tilde G)\rightarrow V(\tilde H)$ such that $f(v)=f(v')$. Hence, for such $\tilde G, v, \tilde H, v'$ it holds that $\mpc_{\tilde G}(v)=\mpc_{\tilde H}(v')\Rightarrow \tilde F(\tilde G)(v)=\tilde F(\tilde H)(v')$. Hence, $\tilde F$ is mp-invariant. Hence, by \Cref{theo:main} there is an R-GNN $N'$ that computes $\tilde F$. Hence, let $N$ be an RNI R-GNN that uses $N'$ as its deterministic R-GNN, then $N$ computes $F$.
\end{proof}

\thmwlinvariant*
\begin{proof}
    The proof imitates the proof of \Cref{theo:main}, reducing WL-invariant functions to R-GNNs with global aggregation, using adapted versions of the intermediate models MPC-GA; MP-LGA; and S-MP-GA, which we will denote by MPC-GA$^w$;MP-LGA$^w$;and S-MP-GA$^w$:
    \begin{itemize}
    \setlength\itemsep{0.1em}
        \item MPC-GA$^w$ differs from MPC-GA in that that it receives the dag representing $\wl_G(v)$ instead of $\mpc_G(v)$
        \item MP-LGA$^w$ differs from MP-LGA in that that in each iteration it has 3 inputs rather than 2 - the third input being the multiset of values of $V(G)\setminus N_G(v)$.
        \item S-MP-GA$^w$ differs from S-MP-GA in that that in each iteration it has 5 inputs rather than 4 - the additional input being the sum of values of $V(G)$.        
    \end{itemize}
    The reduction from MPC-GA$^w$ to MP-LGA$^w$ is a straightforward adaptation of the reduction from MPC-GA to MP-LGA.
    Reducing MP-LGA$^w$ to S-MP-GA$^w$ is similar to reducing MP-LGA to S-MP-GA: It is not difficult to see how to modify the 'receive\_max' procedure (see \Cref{lst:mpgc_to_mpgc_one}) to use a new input 'global\_sum' and, in addition to the current update of the multiset of neighbors values, update a multiset of the values of $V(G)\setminus N_G(v)$, thus collecting the required information for emulating an MP-LGA$^w$ iteration.
    Finally, reducing S-MP-GA$^w$ to R-GNN with global sum-aggregation can be achieved by replicating the processing of the neighbors' sum:
    \begin{itemize}
    \setlength\itemsep{0.1em}
        \item Having a sub-network $F_{B2Q_s}$, similar to $F_{B2Q}$, to translate the received global sum from $\RB$ to $\RQ$.
        \item Having dimensions in $F_M$ to accommodate the global sum input, thus being able to emulate a Turing machine that has the same inputs as an S-MP-GA$^w$.
        \item Having connections between $F_{B2Q_s}$ and other sub-networks, similar to those that $F_{B2Q}$ has.
    \end{itemize}
\end{proof}

\end{document}